\documentclass[acmsmall]{acmart}
\makeatletter
\def\@ACM@checkaffil{% Only warnings
    \if@ACM@instpresent\else
    \ClassWarningNoLine{\@classname}{No institution present for an affiliation}%
    \fi
    \if@ACM@citypresent\else
    \ClassWarningNoLine{\@classname}{No city present for an affiliation}%
    \fi
    \if@ACM@countrypresent\else
        \ClassWarningNoLine{\@classname}{No country present for an affiliation}%
    \fi
}
\makeatother

\usepackage{amsmath}
\usepackage{amsthm}
\usepackage{hyperref}
\usepackage[capitalise]{cleveref}
\usepackage{algorithm}
\usepackage[noend]{algpseudocode}
\usepackage{enumerate}

\renewcommand\footnotetextcopyrightpermission[1]{} % removes footnote with conference information in first column
\newcommand{\sz}{\vec{0}}

\newcommand{\s}{\mathbb{S}}
\newcommand{\pr}[1]{\mathbb{P}\{#1\}}
\newcommand{\Ep}{\mathbb{E}}
\newcommand{\rbar}{\bar{r}}
\newcommand{\rhat}{\hat{r}}
\newcommand{\parahead}[1]{\pink{\textbf{#1~~}}}

\title[Convergence of NPG for a Family of Infinite-State Queueing MDPs]{Convergence of Natural Policy Gradient for a Family of Infinite-State Queueing MDPs}

    \newcommand{\pink}{}

\newtheorem{theorem}{Theorem}%  meant for continuous numbers
\newtheorem{lemma}{Lemma}
\newtheorem{assumption}{Assumption}
\crefname{assumption}{Assumption}{Assumptions}
\makeatletter
\newtheorem*{rep@theorem}{\rep@title}
\newcommand{\newreptheorem}[2]{%
\newenvironment{rep#1}[1]{%
 \def\rep@title{#2 \ref{##1}}%
 \begin{rep@theorem}}%
 {\end{rep@theorem}}}
\makeatother

\newreptheorem{theorem}{Theorem}
\newreptheorem{lemma}{Lemma}

\author{Isaac Grosof}
\email{izzy.grosof@northwestern.edu}
\affiliation{
\institution{Northwestern University}
\department{Industrial Engineering \& Management Science}
\city{Evanston, IL, USA}
}
\author{Siva Theja Maguluri}
\email{siva.theja@gatech.edu}
\affiliation{
\institution{Georgia Institute of Technology}
\department{Industrial \& Systems Engineering}
\city{Atlanta, GA, USA}
}
\author{R. Srikant}
\affiliation{
\institution{University of Illinois, Urbana-Champaign}
\department{Electrical \& Computer Engineering}
\city{Urbana, IL, USA}
}
\email{rsrikant@illinois.edu}

\begin{abstract}
A wide variety of queueing systems can be naturally modeled as infinite-state Markov Decision Processes (MDPs). In the reinforcement learning (RL) context, a variety of algorithms have been developed to learn and optimize these MDPs. At the heart of many popular policy-gradient based learning algorithms, such as natural actor-critic, TRPO, and PPO, lies the Natural Policy Gradient (NPG) policy optimization algorithm. Convergence results for these RL algorithms rest on convergence results for the NPG algorithm. However, all existing results on the convergence of the NPG algorithm are limited to finite-state settings.

We study a general class of queueing MDPs, and prove a $O(1/\sqrt{T})$ convergence rate for the NPG algorithm, if the NPG algorithm is initialized with the MaxWeight policy. This is the first convergence rate bound for the NPG algorithm for a general class of infinite-state average-reward MDPs. Moreover, our result applies to a beyond the queueing setting to any countably-infinite MDP satisfying certain mild structural assumptions, given a sufficiently good initial policy. Key to our result are state-dependent bounds on the relative value function achieved by the iterate policies of the NPG algorithm.
\end{abstract}

\begin{document}

\maketitle
\thispagestyle{empty}
\fancyfoot{}

%\isaac{TODOs: Change framing to emphasize queueing, because pure ML people won't like our assumptions. After motivating with queueing, explain that it's a big deal because no one's done infinite average without special structure, independent interest.}

% Dan Russo's "Global optimality guarantees for policy gradient methods." has now appeared in Operations Research
\section{Introduction}

Policy optimization for queueing systems is a major challenge, both for system operators and for theorists.
While explicit optimal policies may be known in highly simplified settings or in asymptotic regimes, 
for general systems algorithmic optimization is required.
A popular approach is to use optimization techniques from reinforcement learning (RL) to find optimal control policies for these queueing systems,
such as optimal scheduling policies.
For instance, the Natural Policy Gradient (NPG) algorithm lies at the heart of many popular RL algorithms,
such as natural actor-critic, Trust Region Policy Optimization (TRPO) and Proximal Policies Optimization (PPO).
While reinforcement learning techniques are widely used in queueing systems with unbounded queues,
theoretical progress has not kept up with practical usage.
The convergence of NPG in the setting of countably infinite Markov decision problems MDPs,
including the infinite MDPs that arise when optimizing queueing systems, is an open problem. 

The NPG algorithm has been well-studied in recent years, but only in settings with finite state spaces \cite{even-dar_online_2009,murthy_convergence_2023,agarwal_theory_2021}.
Convergence results for the NPG algorithm in these settings have been proven,
and from these results, convergence results for NPG-based reinforcement learning algorithms have also been proven.
The standard progression in this area is to start by studying NPG in the perfect-information, tabular setting, and then build on the ensuing results to analyze settings with learning and with function approximation.

We initiate that progression in the setting of countably-infinite queueing MDPs, or more generally countably-infinite MDPs with certain continuity properties and a single, connected high-reward region of the Markov chain, as we discuss in \cref{sec:assumptions}. These properties are satisfied by a wide variety of MDPs that arise in the setting of queueing theory, as well as many MDPs arising from other settings.

In this setting, we prove that the NPG algorithm converges to an optimal policy, with a concrete bound on the convergence rate.
Our result is the first convergence result for perfect-information NPG in any general countably-infinite setting.

\textbf{Main contributions:}
\begin{itemize}
    \item In \cref{thm:queueing-main}, we prove that in any queueing system in general class of queueing systems known as the Generalized Switch with Static Environment, the NPG algorithm converges to the globally optimal policy, for the mean queue length objective, or more generally the $\alpha$th-moment of queue length for any $\alpha \ge 1$.
    Moreover, we prove an $O(1/\sqrt{T})$ convergence rate bound.
    This result is the first convergence result for the NPG algorithm in any infinite-state queueing setting.
    \item More generally, in \cref{thm:general-main}, we prove that in any countably-infinite average-reward MDP, given a sufficiently good initial policy, the NPG algorithm converges to the optimal policy with an $O(1/\sqrt{T})$ convergence rate bound, given mild structural assumptions on the MDP.
    This is the first such result for any general class of infinite-state average-reward MDPs.
    Prior work in the finite-state setting required strong assumptions on all policies and over all states, as we discuss in \cref{sec:discussion} \cite{murthy_convergence_2023,even-dar_online_2009}.
    \item Methodologically, we have two key novelties in our proof. First, we initialize the NPG algorithm with the MaxWeight policy, rather than with an arbitrary policy, as in previous proofs in the finite state-space setting. The low growth rate of the relative value function of MaxWeight in particular is key to our result. Second, we show that the relative value function only mildly changes between iterations of NPG, allowing us to bound the growth rate of the relative value function of all iterations of the NPG algorithm. Using a state-dependent NPG step size, we adjust for the growing relative value function and prove our convergence result. We discuss our approach in more detail in \cref{sec:discussion,sec:proof-sketch}.
\end{itemize}

This paper focuses on average-reward performance objectives for queueing problems, rather than discounted-reward objectives, as average-reward objectives are most often of importance to queueing practitioners. 
%More generally, this paper focuses on infinite-state average-reward MDPs, rather than discounted-reward MDPs.
While infinite-state discounted-reward MDPs have rarely been explicitly studied,
results on the NPG algorithm in such settings follow from existing finite-state NPG results in the discounted setting,
given the special structure of queueing MDPs.
This is in contrast to infinite-state average-reward results, for which new techniques were needed.
We demonstrate the convergence of NPG for infinite-state discounted-reward MDPs in \cref{sec:discounted},
generalizing a standard finite-state proof.

As a disclaimer, this paper does not consider the practical considerations of function approximation and computational complexity.
This paper focuses on a setting where exact state-action relative values are available for every state,
and studies the exact NPG algorithm, which manipulates and updates a policy with an infinitely long description.
We leave the problems of analyzing the effects of approximate state-action relative value
and analyzing computationally efficient variants of the NPG algorithm to concurrent and future work.
See \cref{sec:concurrent} for our concurrent work on function approximation in the infinite-state setting.
One approach to achieving computational efficiency
is to incorporate a neural-network-based policy approximation, which we leave to future work.

The paper is organized as follows:
\begin{itemize}
    \item \cref{sec:prior-work}: We discuss prior work on the NPG algorithm.
    \item \cref{sec:model}: We define the MDP model, define our queueing model, and define the Natural Policy Gradient algorithm, and define the key properties of queueing MDPs that allow our proof to work.
    \item \cref{sec:results}: We state our convergence result for the NPG algorithm in the setting of countably-infinite average reward queueing MDPs, and discuss the key challenges and a sketch of our proof.
    \item \cref{sec:proof}: We prove our convergence result for the NPG algorithm for queueing MDPs.
    \item \cref{sec:discounted}: In the discounted-reward setting,
    we generalize a standard finite-state convergence proof to the infinite-state setting.
\end{itemize}

\section{Prior work}
\label{sec:prior-work}

We overview prior results on the NPG algorithm in the finite-state setting in \cref{sec:npg-finite},
connections between the NPG algorithm and a variety of reinforcement learning algorithms in \cref{sec:npg-rl},
results on the NPG algorithm in highly-structured infinite-state settings in \cref{sec:npg-structured},
and prior work applying reinforcement learning to queueing problems in \cref{sec:queueing-rl}.
Finally, we discuss concurrent work on a related, but distinct setting in \cref{sec:concurrent}.

\subsection{Natural Policy Gradient with finite state-space}
\label{sec:npg-finite}

The Natural Policy Gradient (NPG) algorithm for MDP optimization
utilizes the Fisher information associated with a policy to choose
a gradient descent direction to update an MDP policy \cite{kakade_natural_2001}.
The NPG algorithm has been shown to have attractive properties,
including global convergence in the tabular setting with finite state space
\cite{agarwal_theory_2021,even-dar_online_2009,murthy_convergence_2023}.

In the discounted-reward tabular setting with finite-state space,
NPG has been shown to converge to the optimal policy at a rate of $O(1/T)$:
The expected discounted reward of the policy after $T$ iterations of NPG
is within $O(1/T)$ of the optimal expected discounted reward \cite{agarwal_theory_2021,geist_theory_2019}.

As we discuss in \cref{sec:discounted}, this result generalizes from the finite-state-space setting with discounted reward to the infinite-state-space setting with discounted reward, given mild assumptions on the structure of the infinite-state MDP,
proving the same $O(1/T)$ convergence result.

In the average-reward tabular setting with finite-state space,
NPG has long been known to converge to the optimal policy at a rate of $O(1/\sqrt{T})$ \cite{even-dar_online_2009}.
Recently, an improved convergence rate result of $O(1/T)$ has been proven for the average-reward finite-state setting
\cite{murthy_convergence_2023}.

However, as we discuss in \cref{sec:discussion},
these average-reward results do not generalize from the finite-state-space setting to the infinite-state-space setting.
Each of the finite-state-space results makes crucial assumptions bounding the behavior of a worst-case state in the MDP.
If the state space is finite, these assumptions are reasonable, but in an infinite-state space setting, the assumptions fail.
Proving convergence in this infinite-state average-reward setting is the focus of this paper.

\subsection{NPG and Reinforcement Learning}
\label{sec:npg-rl}

The Natural Policy Gradient algorithm can be thought of as applying the Mirror Descent framework,
using a Kullback-Leiber divergence penalty to
regularize the basic Policy Gradient algorithm \cite{geist_theory_2019}.

Many reinforcement learning algorithms have been built off of this core idea. The first such algorithm was the Natural Actor-Critic algorithm \cite{peters_natural_2008}, which explicitly builds off of the NPG algorithm,
and which also generalizes previously existing reinforcement learning algorithms such as the original Actor-Critic algorithm \cite{sutton_reinforcement_1998}
and Linear Quadratic Q-Learning \cite{bradtke_adaptive_1994}.

More recent, practically used reinforcement learning algorithms,
such as the Trust Region Policy Optimization (TRPO) \cite{schulman_trust_2015}
and Proximal Policy Optimization (PPO) \cite{schulman_proximal_2017},
also build off the core NPG algorithm.
TRPO replaces the NPG algorithm's KL divergence regularization term
with a KL divergence \emph{constraint}, allowing the algorithm to take larger steps towards an improved policy,
and achieve better empirical performance.
PPO further tweaks the handling of the KL divergence, introducing a clipped surrogate objective based off of the KL divergence.

Further extensions of the PPO algorithm have recently been proposed and empirically studied,
including in the setting of infinite-state average-reward reinforcement learning \cite{dai_queueing_2022}.
These extensions reduce the variance of the value estimation process, thereby improving learning performance.

The theoretical analysis of the NPG algorithm underpins the theoretical motivation for these reinforcement learning algorithms.
We prove the first convergence result for NPG in the infinite-state average-reward setting,
and specifically in the setting of queueing MDPs,
giving theoretical backing for the use of NPG-based reinforcement learning algorithms in this setting.

\subsection{Natural Policy Gradient for specialized settings}
\label{sec:npg-structured}

The policy gradient algorithm has been studied in certain specialized settings
with infinite state-spaces and average-reward objective.
These results rely heavily on the specific details of those settings, and do not generalize beyond those specific contexts.

Fazel et al. \cite{fazel_global_2018} study the Linear Quadratic Regulator (LQR),
an important problem in control theory
which can be thought of as an MDP optimization problem with an uncountably infinite state-space and average reward.
The paper proves that in this setting, the naive policy gradient algorithm converges to the optimal policy,
and the NPG algorithm converges with a faster rate guarantee.
Kunnumkal and Topaloglu \cite{kunnumkal_using_2008} study the base-stock inventory control problem,
and prove that an algorithm using the policy gradient framework achieves convergence to the optimal policy,
with a bound on the convergence rate.

Follow-up study of these settings has recently demonstrated
that these settings exhibit additional structure which allows these convergence results to hold \cite{bhandari_global_2024}.
This additional structure relates to the interplay between policy \emph{gradient} algorithms, which the above results focus on,
and policy \emph{improvement} algorithms.
In particular, it has been shown that if the standard policy improvement algorithm exhibits no suboptimal stationary points,
then policy gradient algorithms will track the standard policy iteration algorithm and thereby converge to the global optimum.
This mechanism is the root cause behind the prior convergence results in the LQR and base-stock settings.

In contrast, such properties do not hold for general infinite-state MDPs,
such as the queueing MDPs we study,
which have much less structure than these previously-studied settings.

\subsection{Queueing and Reinforcement Learning}
\label{sec:queueing-rl}

In this paper, we study MDP optimization for a general class of queueing MDPs.
A variety of papers have studied applying reinforcement learning to queueing problems,
including using learning algorithms which build off of the NPG algorithm.

Dai and Gluzman \cite{dai_queueing_2022} study variance reduction techniques for the PPO algorithm in the context of queueing models,
and demonstrate empirically strong performance, converging towards an optimal policy in several queueing settings.
The theoretical results in the paper focuses on reducing noise that arises from the sampling process, while our results focus on achieving convergence even in a setting where no noise is present at all.

Che et al. \cite{che_dynamic_2023} further employ a differentiable simulation-based modeling technique, allowing additional approximate gradient information to be derived, which is then used to improve the empirical reinforcement learning performance in a queueing context.

Wei et al. \cite{wei_sample_2023} have demonstrated that a ``sample augmentation'' technique can reduce the amount of sample data necessary to converge towards an optimal policy for pseudo-stochastic reinforcement learning settings, including queueing models.
However, the paper focuses on discounted-cost finite-state queueing MDPs, rather than the infinite-state average-reward queueing models which are preferred elsewhere in queueing theory.

Adler and Subramanian \cite{adler_bayesian_2023} study an infinite-state, average-reward queueing MDP optimization setting, where the underlying queueing MDP is parameterized by an unknown parameter $\theta$.
Queueing models with unknown dynamics are used as the motivating models.
The paper assumes that an optimal policy $\pi_\theta^*$ is known for the MDP under any given parameter $\theta$, and focuses on learning the true parameter value $\theta^*$, and on minimizing the regret experienced while learning that parameter.
Our result complements this paper, as we show that the NPG algorithm can be used to find optimal policies for MDPs in a setting where the underlying parameters are exactly known, providing the necessary input $\pi_\theta^*$ for the above result.

In work currently under submission, Chen et al. \cite{chen_primal_2023} study a primal-dual optimization technique for optimizing constrained MDPs,
such as those encountered in queueing problems.
They consider a Langrangian relaxation of the original MDP, and simultaneously optimize the policy and the dual weights of the relaxation.
In the special case of an unconstrained MDP, the underlying primal algorithm they study is the NPG algorithm.

While they consider an infinite-state, average-reward setting, they make a key assumption, \cite[Assumption~3]{chen_primal_2023},
assuming that the state-action relative value function $Q_\pi(s, a)$ is bounded.
This is true in the finite-state average-reward setting, and in the infinite-state discounted-reward setting with bounded reward,
but fails to hold in the general infinite-state average-reward setting which we study.
Correspondingly, in their empirical evaluation, they study queueing MDPs where the queue length has been truncated,
allowing their assumption to hold.
Using that assumption, they prove an $O(1/\sqrt{T})$ convergence rate to the optimal policy.
In contrast, our result is the first to study the NPG algorithm in a setting of unbounded relative value $Q_\pi(s, a)$,
allowing our results to apply to a much more general class of infinite-state, average-reward MDPs,
including queueing models with unbounded queue length.

\subsection{Concurrent Work}
\label{sec:concurrent}

In concurrent work currently under submission, Murthy et al. \cite{murthy_performance_2024}  study a distinct, but related setting using the same framework as in this paper.

That paper also considers a class of infinite-state-space finite-action-space MDPs,
with the objective of maximizing average reward.
The class of MDPs considered in that work is more general in structure than the MDPs considered in this paper,
but the policy space is more restricted.
That paper does not make the structural assumption \cref{ass:reward} and uses a modified version of the connectivity assumption \cref{ass:connected}.
Instead, that paper makes three assumptions:
\begin{enumerate}
    \item For all policies $\pi$ in the policy space $\Pi$, the induced Markov chain is irreducible.
    \item There exists a fixed Lyapunov function $f$ and a fixed finite set $B$ such that for every policy $\pi \in \Pi$,
    the function $f$ and finite set $B$ prove the stability of the underlying Markov chain via the Foster-Lyapunov theorem \cite{foster_stochastic_1953,bremaud_markov_1999},
    in a uniform fashion.\label{item:comp-2}
    \item The finite set $B$ is uniformly connected under all policies $\pi$, similarly to \cref{ass:connected}.
\end{enumerate}
Note that \cref{item:comp-2} in the concurrent paper implies that all policies $\pi \in \Pi$ are stable,
and further implies a uniform lower bound on the average reward of all policies $\pi \in \Pi$.
This paper makes no such assumption.

Furthermore, the concurrent paper studies the noisy NPG algorithm,
where function approximation is used to estimate the state-action relative value $Q_k(s, a)$,
and policy updates are performed using the approximate relative value.
In contrast, our paper only studies the noisy-free NPG algorithm.
Both papers set the step-size $\beta_s$ to depend on the state $s$.

Neither result is a strengthening of the other: This paper examines a general policy space under noise-free NPG updates,
while the Murthy et al. \cite{murthy_performance_2024} paper allows noisy NPG updates but requires a restricted policy space with uniform stability.

\section{Model, Assumptions, and Algorithm}
\label{sec:model}

In this paper, we study the Natural Policy Gradient (NPG) algorithm for MDP optimization, focusing on the MDPs corresponding to a highly general class of queueing models called the Generalized Switch with Static Environment (GSSE). We define general Markov Decision Process notation in \cref{sec:mdp-model}.
We define the NPG algorithm in \cref{sec:npg-alg}.
We define the GSSE class of queueing models in \cref{sec:def-gsse}.

In \cref{sec:results}, we show that the Natural Policy Gradient algorithm converges to an optimal policy for any queueing setting in the GSSE class, when initialized with the well-studied MaxWeight policy, which we define in \cref{sec:mw-def}.
Our analysis of the NPG algorithm generalizes to queueing MDPs beyond the GSSE class,
as well as beyond the queueing setting, to prove convergence for any infinite-state average-reward MDP that satisfies certain structural assumptions, when initialized with a sufficiently good initial policy. We specify these structural assumptions in \cref{sec:assumptions}.

\subsection{Markov Decision Processes}
\label{sec:mdp-model}

We consider Markov Decision Processes (MDPs)
with infinite horizon and
infinite state space $S$.
We assume that for each state $s \in S$,
there exists a finite action space $A_s$,
with a uniform bound on the size of the action space $\vert A_s\vert \le A_{\max}$.
We define the policy space $\Pi$ to be the set of all functions $\pi$ from states to distributions over actions.

The environment is captured by a transition function
$\mathbb{P}$ which maps a state-action pair $(s, a)$ to a distribution over states $s' \in S$,
denoted by
$\pr{s' \mid s, a}$.
For convenience, we will write $\pr{s' \mid s, \pi}$ to denote the probability
of transition from $s$ to $s'$ under policy $\pi$:
\begin{align*}
    \mathbb{P}_\pi\{s' \mid s\} := \pr{s' \mid s, \pi} := \sum\nolimits_{a \in A_s} \pi(a \mid s) \pr{s'\mid s,a}.    
\end{align*}

The reward is given by a reward function
$r(s, a)$ which maps a state-action pair to a real-valued reward.
We likewise write $r(s, \pi)$ to denote the expected single-step reward associated with policy $\pi$ and state $s$:
\begin{align*}
    r(s, \pi) = \Ep_{a \sim \pi(\cdot \mid s)}[r(s, a)]
\end{align*}
We define $r_{\max}(s)$ to be the maximum reward achievable in state $s$,
$r_{\max}(s) := \max_a r(s, a)$.
Note that we focus on reward-maximization rather than cost-minimization.
Our primary reward function is the negation of the total queue length.

If $\pi$ gives rise to a stable Markov chain, let $J_\pi$ denote the average reward associated with policy $\pi$,
defined as
\begin{align*}
    J_\pi = \lim_{T \to \infty} \frac{1}{T} \Ep_\pi \left[\sum\nolimits_{i=0}^{T-1} r(s_i, \pi) \right],
\end{align*}
where $s_i$ is the state at time $i$,
and where the expectation $\Ep_\pi$ is taken with respect to the transition probability $\mathbb{P}_\pi$.

Let $d_\pi$ denote the stationary distribution over $S$
under the policy $\pi$.
We can express $J_\pi$ as
\begin{align*}
    J_\pi = \Ep_{s \sim d_\pi} [ r(s, \pi)].
\end{align*}

The state relative value function $V_\pi(s)$ is defined, up to an additive constant $C$, to be the
additive transient effect of the initial state on the total reward:
\begin{align*}
    V_\pi(s) = C + \lim_{T \to \infty} \left( \Ep_\pi \left[\sum\nolimits_{i=0}^{T-1} r(s_i, \pi) \mid s_0=s\right] - J_\pi T \right)
\end{align*}

This is also the solution, up to an additive constant, of the Poisson equation:
\begin{align}
    \label{eq:poisson-v}
    J_\pi + V_\pi(s) = r(s, \pi) + \Ep_{s' \sim\mathbb{P}_\pi(s)} [V_\pi(s')]
\end{align}

We also define the state-action relative value function $Q_\pi(s, a)$
and its associated Poisson equation:
\begin{align}
    \nonumber
    Q_\pi(s, a) &= C + \lim_{T \to \infty} \left(\Ep_\pi \left[\sum\nolimits_{i=0}^{T-1} r(s_i, \pi) \mid s_0=s,a_0=a\right] - J_\pi T\right) \\
    \label{eq:poisson-q}
    J_\pi + Q_\pi(s, a) &= r(s, a) + \Ep_{s' \sim\mathbb{P}_a(s)} [Q_\pi(s', \pi)].
\end{align}
To uniquely specify the additive constant $C$, we will adopt the convention that
$V_\pi(\sz) = 0$, for a specially identified state $\sz$ defined in \cref{ass:reward}.
In the queueing setting discussed in \cref{sec:def-gsse}, $\sz$ is the state with no jobs in the system.

Often, we are interested in the case where the action $a$ is sampled from a policy $\pi$'s action distribution $\pi(a \mid s)$. We define corresponding notation:
\begin{align*}
    Q_\pi(s, \pi') := \Ep_{a \sim \pi'(\cdot \mid s)}[Q_\pi(s, a)]
\end{align*}

Let $\tau_\pi(s)$ denote the expected time under policy $\pi$ to hit the specially identified state $\sz$,
starting from state $s$. 

Let $\pi_0, \pi_1, \ldots, \pi_k, \ldots$ denote the iterate policies of the NPG algorithm defined in \cref{sec:npg-alg}.
For concision, we will write $J_k, \tau_k, Q_k,$ etc. as shorthand for $J_{\pi_k}, \tau_{\pi_k}, Q_{\pi_k}$, etc., to denote functions of the iterate policy $\pi_k$.

\subsection{NPG Algorithm}
\label{sec:npg-alg}

This paper studies the Natural Policy Gradient algorithm, given in \cref{alg:npg}.

\begin{algorithm}
\caption{The Natural Policy Gradient algorithm}
\label{alg:npg}
\begin{algorithmic}
\State \textbf{Initialize:}  A learning rate function, mapping states $s$ to learning rates $\beta_s$.
\State \textbf{Initialize:}  An initial policy $\pi_0$.
\For {each iteration $k = 0$ to $T-1$}
  \For {all state-action pairs $s, a$}
  \State Compute the value function $Q_k(s, a)$.
  \State Compute the weighted update $\pi_k(a \mid s) \beta_s^{Q_k(s, a)}$.
  \State Set the new policy probability $\pi_{k+1}(a \mid s) = \pi_k(a \mid s) \beta_s^{Q_k(s, a)} / Z_{s,k},$
  \State where $Z_{s,k} = \sum_{a'} \pi_k(a' \mid s) \beta_s^{Q_k(s, a')}$.
  \EndFor
\EndFor
\end{algorithmic}
\end{algorithm}
Note that the NPG algorithm is an MDP optimization algorithm, rather than a learning algorithm.
In particular, we assume that the relative value function $Q_\pi(s, a)$ can be exactly computed, rather than approximated, and we do not consider the practical implementation of the algorithm, leaving the problem of making the algorithm computationally efficient to future work.

\subsection{Definition of Generalized Switch with Static Environment}
\label{sec:def-gsse}

The Generalized Switch with Static Environment (GSSE) is a queueing model with $n$ classes of jobs,
and the state of the system is a size-$n$ vector $q_\pi(t)=\{q^{(i)}_\pi(t)\}$
specifying the number of jobs of each class present in the system under policy $\pi$.
This model is a special case of the Generalized Switch model \cite{stolyar_maxweight_2004}.
Let $q_\pi = \{q^{(i)}\}$ denote the corresponding stationary random variable.
Note that we use lower case $q$ to denote queue lengths and upper case $Q$ to denote state-action relative value.

The system evolves in discrete time:
At each time step, $v_i$ jobs of class $i$ arrive,
where $v_i$ is a bounded, i.i.d. random variable.
Jobs of each class arrive independently.

There are $m$ service options.
At each time step, prior to new jobs arriving, the scheduling policy selects a service option $j$.
If the policy selects service option $j$,
then $w_i^j$ jobs of class $i$ complete,
where $w_i^j$ is a bounded, i.i.d. random variable.
Jobs of each class complete independently.
If on some time step $t$, $w_i^j(t)$ exceeds $q^{(i)}(t)$, the number of jobs of class $i$ present in the system,
then all of the class $i$ jobs complete.

We use the notation $v, w$ instead of the more common $a, s$ to avoid collision with the action and state notation defined in \cref{sec:model}.

Let $\ell$ be the maximum number of jobs of any class that can arrive or depart in one time slot: The maximum over all $v_i$ and all $w_i^j$.
Let $\lambda_i = E[v_i]$ be the average arrival rate of class $i$ jobs, and let $\mu_i^j = E[w_i^j]$ be the average completion rate of class $i$
jobs under service option $j$.

Our primary reward function $r(s, \cdot)$ of state $s=\{q_i\}$ under any action
is the negative total queue length
$r(s, \cdot) = -\sum_i q_i = - \vert q \vert_1$.
We will also consider the setting where the reward is the negative $\alpha$-th moment of the total queue length,
$-(\sum_i q_i)^\alpha$, for a generic $\alpha \ge 1$.
Note that the specially-identified maximum reward state $\sz$ is the empty state, where $\forall i, q_i = 0$.

We make certain non-triviality assumptions to ensure that the state space is not disconnected under any scheduling policy, namely:
\begin{itemize}
    \item For each class $i$, it is possible for no class $i$ jobs to arrive ($P(v_i = 0) > 0$).
    \item For each class $i$, it is possible for a class $i$ job to arrive ($P(v_i > 0) > 0$).
    \item For each pair $(i, j)$, it is possible for more jobs arrive than depart ($P(v_i > w_i^j) > 0$).
    \item For each pair $(i, j)$, it is possible for equally many jobs to arrive as depart ($P(v_i = w_i^j) > 0$).
\end{itemize}
Less restrictive assumptions are possible -- we make these assumptions for simplicity.

Finally, we assume that the scheduling policy $\pi$ is non-idling:
If there are any jobs in the system, we assume that the policy selects a service option $j$ with a nonzero chance of completing a job.
Subject to that restriction, the scheduling policy may be arbitrary.

More specifically, we define our action space $A_s$ in a given state $s=\{q_i\}$ to be the set of non-idling service options $j$
for the given state $s$, for any state $s$ which is not the empty state:
\begin{align*}
    A_{\{q_i\}} := \{j \mid \exists i: q_i > 0 \land P(w_i^j > 0) > 0\}, \forall \{q_i\} \neq \sz
\end{align*}
As a special case, if $\{q_i\}$ is the empty state $\sz$, any service option may be used: idling is unavoidable.
The uniform bound on the size of the action space is satisfied with $A_{\max} = m$.

Many commonly-studied queueing models fall within the GSSE framework,
with appropriate choices of arrival distributions and service options.
For instance,
the input-queued switch \cite{mekkittikul_starvation_1996},
the multiserver-job model \cite{grosof_wcfs_2022},
and parallel server systems \cite{harrison_heavy_1999} such as the N-system \cite{harrison_heavy_1998},
all fit within the GSSE framework.

Note that the assumption of a static environment can be relaxed.
The static environment assumption is only used to prove drift properties of the MaxWeight policy in \cref{lem:bound-mw},
which can be proven, with more effort, from more relaxed conditions on the environment.

\subsection{MaxWeight}
\label{sec:mw-def}

An important scheduling policy in the Generalized Switch model is the MaxWeight policy \cite{tassiulas_stability_1990},
which chooses the service option which maximizes the inner product of the queue lengths and the service rates:
\begin{align}
\label{eq:maxweight-def}
    \text{MaxWeight}(\{q_i\}) := \arg\max_j \sum\nolimits_i q_i \mu^j_i
\end{align}
Note that MaxWeight always selects a non-idling service option $j$ for any state $\{q_i\} \neq \sz$.
This service option $j$ thus lies within our action space $A_{\{q_i\}}$.

MaxWeight is known to have many advantageous properties in the \emph{heavy traffic} limit. The heavy traffic limit is the limit in which the arrival rates approach the boundary of the capacity region. In the GSSE setting, the capacity region is the convex hull of the available service rates $\mu_i^j$.

In particular, MaxWeight is known to be throughput optimal, meaning that it keeps the system stable for all arrival rates within the capacity region \cite{stolyar_maxweight_2004,hurtado-lange_heavy_2022}.
Moreover, MaxWeight asymptotically minimizes the total queue length $\sum_i q_i$ in the heavy traffic limit \cite{stolyar_maxweight_2004} in our discrete-time setting.

Outside of the heavy-traffic limit, in non-asymptotic regimes, it is known that MaxWeight can be significantly outperformed,
especially by methods based on MDP optimization.
For instance, in the N-system, a threshold-based policy is known to significantly outperform MaxWeight at nonasymptotic arrival rates \cite{bell_dynamic_2001},
and optimization-based methods are needed to find the optimal policy in the non-asymptotic arrival-rate regime \cite{dai_queueing_2022}.

In this paper, we use the MaxWeight policy to initialize our NPG algorithm as $\pi_0$.

\subsection{Assumptions}
\label{sec:assumptions}

The goal of this paper is to study queueing MDPs.
We therefore define a general class of MDPs which include a wide variety of queueing MDPs.
The class of MDPs studied in this paper are MDPs which satisfy two assumptions
on the structure of our MDP: One on the rewards in the MDP,
and one on the connectedness of high-reward states.
We verify these assumptions for GSSE queueing MDPs in \cref{sec:queueing-main}.

First, we make mild assumptions on the reward structure of the MDP, which are straightforwardly satisfied by a wide variety of queueing MDPs:
\begin{assumption}[Reward structure] \label{ass:reward} \mbox{}
    \begin{enumerate}
        \item[(a)] We assume that the MDP has bounded positive reward, though it may have unlimited negative reward (e.g. unlimited cost).
        Specifically, we assume that $c_{\max} := \sup_{s, a} r(s, a) < \infty$.
    \end{enumerate}
    It will be useful in some cases to normalize this upper bound to 0. Let us define the reduced reward $\rhat(s, a) := r(s, a) - c_{\max}$.
    \begin{enumerate}
        \item[(b)] We assume that there are finitely many high-reward states. Specifically, for any $z$, we assume that there are finitely many states $s$ such that $r_{\max}(s)\ge z$.
    \end{enumerate}
    From (a) and (b), it follows that there must exist a state $s^*$ which achieves the maximum reward $r_{\max}(s^*) = c_{\max}$.
    Let $\sz$ be a specific maximum-reward state.
    \begin{enumerate}
        \item[(c)] We assume the reward $\hat{r}(s, a)$ is not overwhelmingly dominated by the action $a$, as opposed to the state $s$.
        Specifically, we assume that there exist constants $R_1, R_2 \ge 0$ such that
    for all states~$s$ and all actions $a$, $\rhat_{\max}(s) - \rhat(s, a) \le R_1 \rhat_{\max}(s)^2 + R_2$.
        \item[(d)] We assume that the reward $r_{\max}(s)$ does not change very quickly between neighboring states.
        Specifically, we assume that there exist constants $R_3 \ge 1, R_4 \ge 0$,
        such that, if $s, s'$ are a pair of states such that $\pr{s \mid s, a}>0$ for some action $a$,
        then $\rhat_{\max}(s') \ge R_3 \rhat_{\max}(s) - R_4$.
    \end{enumerate}
\end{assumption}
In the queueing setting, reward is $r(s, a) = -\vert q \vert_1$, $c_{\max} = 0$, and $\sz$ is the empty-queue state.

Second, we assume that the high-reward states are uniformly connected, under an arbitrary policy. For a queueing MDP, this states that it's possible to move from any short-queue state to any other short-queue state in a uniformly bounded number of steps under an arbitrary policy.
\begin{assumption}[Uniform connectedness of high-reward states]
    \label{ass:connected}
    Given a reward threshold $z$,
    we assume that there exists a number of steps $x_z$
    and a probability $p_z > 0$
    such that for all states $s, s'$ where $r_{\max}(s) \ge z$,
    $r_{\max}(s') \ge z$,
    and for all policies $\pi$,
    the probability that
    the MDP initialized at state $s$
    and transitioning under policy $\pi$
    reaches state $s'$ in at most $x_z$ steps
    is at least $p_z$.
\end{assumption}

Note that for \cref{ass:connected} to hold, \cref{ass:reward}(b) is a necessary prerequisite assumption.
Uniform connectedness of high-reward states is only possible if there are finitely many high-reward states.

\section{Results}
\label{sec:results}

We prove the first convergence result for the Natural Policy Gradient (NPG) algorithm in the setting of unbounded queues.
In particular, we show that for any queueing model in the Generalized Switch with Static Environment (GSSE) class of models, the NPG algorithm rapidly converges to the scheduling policy with minimal mean queue length.
The GSSE setting is a broad, natural family of queueing systems, which includes as special cases such important queueing systems as the $n \times n$ switch, the N-system, and the multiserver-job system.

To achieve this rapid convergence, we initialize the NPG algorithm with the MaxWeight scheduling policy, a well-studied policy that is known to achieve optimal stability region, but which does not achieve optimal or near-optimal mean queue length except in certain asymptotic regimes.
We also select a specific learning rate parameterization $\beta_s$, whose properties are vital to our convergence result.

We demonstrate that after $T$ iterations, the $T$th NPG iterate will achieve a mean queue length within $O(1/\sqrt{T})$ of the optimal policy.

\begin{theorem}
    \label{thm:queueing-main}
    For any GSSE queueing model with the objective of minimizing mean queue length,
    the NPG algorithm with initial policy $\pi_0 = MaxWeight$
    and with learning rate parameterization given in \cref{thm:general-main}
    achieves the following mean queue length:
    \begin{align}
        E[\vert q_{\pi_T} \vert_1] \le E[\vert q_{\pi^*} \vert_1] + \frac{c_*}{\sqrt{T}},
    \end{align}
    where $\pi_T$ is the $T$th iterate of the NPG algorithm,
    $\pi^*$ is the policy with minimal mean queue length,
    and $c_*$ is a constant depending on the specific GSSE model. \\
    This result generalizes to the $\alpha$th moment of queue length.
\end{theorem}
\proof[Proof deferred to \cref{sec:queueing-main}.]{\,}

Beyond the GSSE setting, our approach holds for a more general family of infinite-state average-reward MDPs,
including both queueing systems outside the GSSE family of models, as well as a wide variety of non-queueing systems with countably infinite state spaces.
This is the first NPG convergence result for any general class of infinite-state average-reward MDPs.

We now state our result in full generality:
\begin{theorem}
    \label{thm:general-main}
    For any average-reward MDP satisfying \cref{ass:reward,ass:connected},
    given an initial policy $\pi_0$ such that
    there exist constants $c_0 > 0, c_1 \ge 0$ such that
    \begin{align}
        \label{eq:relative-value-assumption}
        V_0(s) \ge -c_0 \rhat_{\max}(s)^2 - c_1,
    \end{align}
    the NPG algorithm with learning rate parameterization $\beta_s$ given in \eqref{eq:beta-s-main}
    achieves the convergence rate
    \begin{align*}
        J_* - J_T \le \frac{c_*}{\sqrt{T}},
    \end{align*}
    where $c_*$
    is a constant depending on the MDP parameters and on $c_0$ and $c_1$.
\end{theorem}
\proof[Proof deferred to \cref{sec:proof}.]{\,}

We first prove \cref{thm:general-main}, our more general result,
and then apply it to the GSSE queueing setting to prove \cref{thm:queueing-main}.

In \cref{sec:discussion}, we discuss the key challenges which we needed to overcome to prove \cref{thm:queueing-main} and \cref{thm:general-main}.
We discuss our approach in \cref{sec:proof-sketch},
which centers on bounding the growth rate of the relative value function $V_k(s)$ under the NPG iterates $\pi_k$.

Our quadratic-growth-rate assumption in \eqref{eq:relative-value-assumption} on the initial policy
is the most general assumption that suffices for our proof technique.
Intuitively, a key step in the proof of \cref{thm:general-main}
depends on proving that the expected square-root of the relative value of each NPG iterate $V_k(s)$ is finite,
where the state $s$ is distributed according to the stationary distribution of the optimal policy $\pi^*$: $E_{s \sim \pi^*}[\sqrt{V_k(s)}] < \infty$.
Using the quadratic-growth-rate assumption in \eqref{eq:relative-value-assumption},
we can prove that $V_k(s)$ also grows at most quadratically for each NPG iterate,
and thereby bound the expectation in comparison to the mean reward of the optimal policy,
which is known to be small.
Without \eqref{eq:relative-value-assumption}, or with a weaker assumption,
we would have no way to prove that this expectation is finite.
See \eqref{eq:using-main-assumption} and the accompanying discussion in the proof of \cref{thm:general-main} for more detail.

\subsection{Challenges}
\label{sec:discussion}

We seek to prove a convergence rate result for the NPG algorithm in a setting of unbounded queues,
which is an infinite-state-space, average-reward setting.
A natural approach would be to try to generalize finite-state average-reward results on the NPG algorithm
to the infinite-reward setting.
There are two relevant approaches to consider:
the approach of Even-Dar et al. \cite{even-dar_online_2009}, and of Murthy and Srikant \cite{murthy_convergence_2023}.
Unfortunately, each of these approaches makes crucial use of assumptions that are only plausible for MDPs with finite state spaces.

With respect to Even-Dar et al. \cite{even-dar_online_2009},
combining that paper's Algorithm 4 (MDP Experts) with its Algorithm 1 (Weighted Majority)
results in exactly the Natural Policy Gradient algorithm,
and the paper's Theorem~4.1 proves that NPG converges to the optimal policy in the finite-state average-reward setting,
with convergence rate $O(1/\sqrt{T})$.
To prove this theorem, the paper assumes that there exists some finite mixing time $\tau$ for all policies $\pi$.
Intuitively, this assumption states that for every policy
and every initial system state,
the system evolves to a state distribution
near the policy's stationary distribution in at most $O(\tau)$ steps.
In a finite-state MDP, this is a reasonable assumption: For reasonable policies, $\tau$ might be exponential in $\vert S \vert$ at worst.
In contrast, in a queueing system with unbounded queue length and bounded completion rate, this assumption is plainly false.
No policy can achieve a finite mixing time, much less all policies.

Murthy and Srikant \cite{murthy_convergence_2023} prove that the NPG algorithm converges to the optimal policy in the finite-state average-reward setting at rate $O(1/T)$.
To do so, the paper assumes that there exists some uniform lower bound $\Delta$ on the relative state probability of
an arbitrary policy and of the optimal policy, in stationarity:
\begin{align*}
    \Delta = \inf_{\pi, s} \frac{d_\pi(s)}{d_*(s)} > 0
\end{align*}
In a finite-state MDP, this is a reasonable assumption: As long as $d_\pi$ places nonzero probability on every state $s$,
the assumption will be satisfied.
In contrast, in an infinite-state MDP, $d_\pi(s)/d_*(s)$ will approach zero as we look at more and more states $s$, for almost all policies $\pi$.
Restricting to a class of policies $\pi$ where $d_\pi(s)/d_*(s)$ remains bounded away from zero would require advance knowledge of 
the optimal policy $\pi^*$.

Another natural approach would be to try to use results from the discounted-reward setting to prove results in the average-reward setting.
A standard result states that a policy $\pi$'s average reward $J_\pi$ can be related to its discounted reward $V_{\eta,\alpha}^\pi$ via the formula $J_\pi = \lim_{\alpha \to 1}(1-\alpha)V_{\eta,\alpha}^\pi,$
where $V_{\eta,\alpha}^\pi$ denotes a policy $\pi$'s expected reward, starting from distribution $\eta$, with discount factor $\alpha$ \cite{bertsekas_dynamic_2012}.
Unfortunately, existing convergence-rate bounds on NPG in the discounted-reward setting, such as \cite[Theorem 16]{agarwal_theory_2021}, have a $\theta(\frac{1}{(1-\alpha)^2})$
dependency on $\alpha$, so discounted-reward results do not prove that the NPG algorithm converges in the average-reward setting,
much less bound its convergence rate.

These proof approaches do not straightforwardly generalize to our setting of unbounded queues.

\subsection{Proof sketch}
\label{sec:proof-sketch}

Delving deeper into the Even-Dar et al. \cite{even-dar_online_2009} approach,
their key insight is that one can think of the MDP optimization problem as consisting of many instances of the expert advice problem 
\cite{cesa-bianchi_expert_1997},
and think of the NPG algorithm as running a separate instance
    of the weighted majority algorithm for each state $s$,
    where the reward function at time step $k$ is $Q_k(s, a)$.

    In the expert advice problem,
    an agent has a set of possible actions,
    each with a secret reward.
    After an action $a_k$ is chosen on time-step $k$, the reward function $\widetilde{r}_k(\cdot)$ is revealed to the agent.
    The goal of the expert advice problem is to choose a sequence of actions $\{a_k\}$
    whose total reward is close to that of the optimal single action, $a_*$.

    The weighted majority algorithm maintains a distribution $\pi$ over actions,
    sampling an action at random from its distribution at each time step.
    At each time step, the action distribution is updated according to the rule:
    \begin{align*}
        \pi_{k+1}(a) = \beta^{\widetilde{r}_k(a)}/Z_k, \text{ where } Z_k = \sum\nolimits_{a'} \pi_k(a')  \beta^{\widetilde{r}_k(a)}.
    \end{align*}

    \cite[Theorem 4.4.3]{cesa-bianchi_expert_1997}, restated as \cref{lem:weighted-majority}, states that, as long as the reward $\widetilde{r}_k(a)$ is bounded,
    there exists a choice of $\beta$ such that the weighted majority algorithm achieves $O(\sqrt{T})$
    regret when compared to the optimal action.

Thus, to bound the convergence rate of the NPG algorithm,
it is crucial to bound the relative value function $V_\pi(s)$, and thereby bound the state-action relative value $Q_k(s, a)$.
In the finite-state setting, Even-Dar et al. \cite{even-dar_online_2009} use their bounded-mixing time assumption to
prove a universal bound on $V_\pi(s)$ over all policies $\pi$, states $s$, and actions $a$,
which allows the NPG convergence rate proof to be completed.

In our infinite-state setting, we also prove bounds on $V_\pi(s)$,
but we prove policy- and state-dependent bounds.
Specifically, we prove strong enough bounds to complete the proof in a similar fashion to Even-Dar et al. \cite{even-dar_online_2009},
transferring bounds on the relative value function to bounds on the convergence rate.
These relative-value bounds form the key novelty of this paper.

We now outline our lemmas, which primarily focus on bounding $V_\pi(s)$, towards our goal of proving \cref{thm:general-main}, our generalized convergence rate bound,
and ultimately proving \cref{thm:queueing-main}, our convergence rate bound for GSSE queues specifically.

\parahead{Step 0: Prove that each iterate $\pi_k$ is stable}
It is vital for our result that at every NPG iteration $k$,
the current policy $\pi_k$ is stable.
This is not trivial,
as there exist unstable policies $\pi$ within our space of eligible policies.

In \cref{lem:stable}, we inductively prove that each policy $\pi_k$ is stable by using the relative value function $V_{k-1}(s)$ of the previous policy $\pi_{k-1}$ as a Lyapunov function.

\parahead{Step 1: Bounding change in relative value under NPG update}
The presence of unstable policies $\pi$ shows that for general policies $\pi$,
$V_\pi(s)$ may be unbounded.
However, we don't need to bound $V_\pi(s)$ over all policies $\pi$.
Instead, we focus on the specific policies $\pi_k$ visited by the NPG algorithm,
and bound the relative value functions $V_k(s)$ of only those policies.

We start with \cref{lem:rel-policy} by bounding how much worse $V_{k+1}(s)$ can be than $V_k(s)$,
relative to the policy's hitting time $\tau_k(s)$ from arbitrary states $s$ to the special state $\sz$.
This lemma builds on \cref{lem:monotonic-q},
a standard monotonicity result for the NPG policy.

\parahead{Step 2: Bounding high-reward states}
Next, in \cref{lem:uniform-pr}, we bound the time $\tau_\pi(s)$ for the system to hit $\sz$,
starting from a high-reward state $s$ under
policies $\pi$ with high average reward $J_\pi$.
Here, we make use of the fact that the NPG algorithm is known to monotonically increase the average reward at each iteration (\cref{lem:monotonic-j}),
so this result applies to all iterates of the algorithm.

\parahead{Step 3: Bounding all states}
Putting it all together, in \cref{lem:bound-vk}, we bound the relative value $V_k(s)$ of the iterate policies
in comparison to the relative value $V_0(s)$ of the initial policy,
for arbitrary states $s$.
Thus, if the initial policy has a well-behaved relative value function,
then each iterate policy will also have a well-behaved relative value function,
and we can prove fast convergence.

\parahead{Step 4: Main general result, using the quadratic assumption}
In \cref{thm:general-main},
we specifically assume that the initial policy's relative value function $V_0(s)$ grows at most quadratically relative to the reward $r(s, a)$,
an assumption that we show in \cref{lem:bound-mw} is satisfied by the MaxWeight policy in the setting of queueing MDPs.

As a result, we now have a bound on the relative value function $V_k(s)$ for each NPG iterate $\pi_k$
which depends on the state $s$ -- the bound is not uniform over all states.
We specify this bound in \cref{lem:bound-vk-state}.
Correspondingly, we set our learning rate $\beta$ as a function of the state $s$,
based on the bound on the relative value function that we are able to prove in that particular state.
In states with longer queues (more negative reward), our bound on $V_\pi(s)$ is weaker,
so we use a slower learning rate, to improve convergence.

Now, with our state-dependent bounds on the relative value function for the NPG iterate policies,
we are ready to employ the approach outlined at the beginning of this section:
Thinking of the MDP optimization problem as many instances of the expert advice problem,
and thinking of the NPG algorithm has many instances of the weighted majority algorithm for that problem.

In \cref{thm:general-main}, we prove that by starting with a initial policy $\pi_0$
whose relative value function $V_0(s)$ grows at most quadratically with respect to reward $r(s, a)$,
and by selecting the right $\beta_s$ function,
the NPG algorithm is guaranteed to achieve a $O(1/\sqrt{T})$ convergence rate.
Our quadratic assumption on the initial policy is exactly strong enough to prove this convergence rate.

\parahead{Step 5: Main queueing result}

Finally, in \cref{thm:queueing-main}, we prove that every queueing system in the GSSE family of queues satisfies the requirements for \cref{thm:queueing-main} to apply,
and that the MaxWeight policy's relative value function $V_{MW}(s)$ grows at most quadratically with respect to total queue length. As a result, we show that the NPG algorithm is guaranteed to achieve a $O(1/\sqrt{T})$ convergence rate for any queueing model in the GSSE family.

\section{Proofs}
\label{sec:proof}

We start by stating background lemmas from the literature in \cref{sec:background}.
Then we proceed with our proof:
\begin{enumerate}
    \item In \cref{lem:stable}, we show that all NPG iterates $\pi_k$ are stable.
    \item In \cref{lem:rel-policy}, we bound the change in relative value between two policies which are consecutive iterates of the NPG algorithm.
    \item In \cref{lem:uniform-pr}, we bound the hitting time $\tau_k(s)$ for high-reward states $s$ under high-average-reward policies, such as the policies which are iterates of the NPG algorithm.
    \item In \cref{lem:bound-vk}, we bound the relative value of all NPG iterates in all states $s$, in comparison to the initial policy $\pi_0$.
    \item In \cref{sec:main-proof}, we prove our main result stated in full generality, \cref{thm:general-main}, making use of our quadratic-relative-value assumption on our initial policy $\pi_0$.
    \item In \cref{sec:bound-mw}, we prove that the MaxWeight policy achieves a quadratically-growing relative value function for any queueing system in the Generalized Switch with Static Environment (GSSE) family of queues.
    \item Finally, in \cref{sec:queueing-main}, we prove \cref{thm:queueing-main}, our main result stated specifically for GSSE queues.
\end{enumerate}

\subsection{Background lemmas}
\label{sec:background}

We start by restating results from the literature.
Note that \cref{lem:monotonic-q,lem:monotonic-j} were each originally proven in a setting with fixed $\beta$, but the proofs still hold unchanged in our setting with variable $\beta_s$.

First, we state two monotonicity results for the NPG algorithm,
both proven in \cite[Lemma 2]{murthy_convergence_2023}.

\begin{lemma}
    \label{lem:monotonic-q}
NPG iterate policy $\pi_{k+1}$ improves upon policy $\pi_k$ relative to $\pi_k$'s relative-value function $Q_k$:
    \begin{align*}
        Q_k(s, \pi_{k+1}) \ge Q_k(s, \pi_k) = V_k(s)
    \end{align*}
\end{lemma}
\begin{proof}
This result is proven for the case of non-state-dependent learning rate $\beta$ in \cite[Lemma 2]{murthy_convergence_2023}.
We demonstrate that the proof transfers to the state-dependent case in \cref{app:background}.
The proof holds for infinite MDPs, and does not require $\pi_{k+1}$ to be a stable policy.
\end{proof}

Next, we state the Performance Difference Lemma,
\cref{lem:perf-diff}, which relates the difference in average reward between two policies to their relative value functions:
\begin{lemma}{Performance Difference Lemma: \cite[(10)]{cao_single_1999}}
    \label{lem:perf-diff}
        For any pair of stable policies $\pi, \pi'$,
    \begin{align*}
        J_\pi - J_{\pi'} = E_{s \sim d_\pi}[Q_{\pi'}(s, \pi) - V_{\pi'}(s)].
    \end{align*}
\end{lemma}
In particular, \cref{lem:perf-diff} holds in our infinite-state setting.
The proof of \cref{lem:perf-diff} given by Cao \cite{cao_single_1999} requires only that the underlying Markov chains are ergodic,
ensuring that the relevant quantities are well-defined.

\begin{lemma}
    \label{lem:monotonic-j}
    If NPG iterate policies $\pi_k$ and $\pi_{k+1}$  are stable, then average reward increases monotonically: $J_{k+1} \ge J_k$.
\end{lemma}
\begin{proof}
    \cref{lem:monotonic-j} follows immediately from \cref{lem:monotonic-q} and \cref{lem:perf-diff}.
\end{proof}

Finally, we state a result on the weighted majority algorithm for the expert advice problem. The problem and algorithm are described in \cref{sec:proof-sketch}.
Note that this paper operates in a reward maximization framework rather than the loss minimization framework of \cite{cesa-bianchi_expert_1997}, so we invert $g(z)$ relative to that paper.
\begin{lemma}{\cite[Theorem 4.4.3]{cesa-bianchi_expert_1997}}
    \label{lem:weighted-majority}
    Consider an instance of the expert advice problem, where for any step $k$ and any pair of actions $a, a' \in  A$,
    $M$ is an upper bound on the value of $\widetilde{r}_k(a) - r_k(a')$.
    By selecting
    \begin{align}
        \label{eq:beta-s}
        \beta = g \left(\sqrt{\frac{\ln \vert A\vert }{T M}}\right),
        \text{where } g(z) = 1 + 2z + z^2/\ln 2,
    \end{align}
    the weighted majority algorithm achieves the following regret guarantee:
    \begin{align*}
        \sum\nolimits_{k=1}^T r_k(a_*) - r_k(a_k) \le \sqrt{T M \ln \vert A\vert } + \log_2(\vert A\vert )/2 .
    \end{align*}
\end{lemma}
\subsection{Step 0: Stability}

We start by showing that each NPG iterate is stable, via induction over the iterations.
\begin{lemma}
    \label{lem:stable}
    For any Markov chain satisfying \cref{ass:reward},
    if the NPG algorithm is initialized with a stable policy $\pi_0$,
    then all NPG iterates $\pi_k$ are stable.
\end{lemma}
\begin{proof}
    We will proceed by induction. We must show that if an NPG iterate $\pi_k$ is stable, then the next NPG iterate $\pi_{k+1}$ is stable.

    We will use the function $b(s) := -V_k(s)$ as a Lyapunov function for the policy $\pi_{k+1}$,
    and apply the Foster-Lyapunov theorem \cite{foster_stochastic_1953,bremaud_markov_1999}.
    To show that $\pi_{k+1}$ is a stable policy with $b(s)$ as our Lyapunov function, we must show that
    \begin{itemize}
        \item $b(s)$ is lower bounded by a constant for all $s$,
        \item There exists a constant $C$ such that for all $s$, the expected one-step update of $b(s)$ is upper bounded:
        \begin{align*}
            \forall s, \, E_{s' \sim P_{k+1}(s)}[b(s')] - b(s) \le C,
        \end{align*}
        \item There exists a finite set $S$ and positive constant $\epsilon > 0$ such that the expected one-step update of $b(s)$ is bounded below zero:
        \begin{align*}
            \forall s \not\in S, \, E_{s' \sim P_{k+1}(s)}[b(s')] - b(s) \le -\epsilon.
        \end{align*}
    \end{itemize}
    Translating these requirements on $b(s) = -V_k(s)$ into requirements on $V_k(s)$,
    we must:
    \begin{itemize}
        \item \textit{upper bound} $V_k(s)$ over all states $s$,
        \item \textit{lower bound} the expected one state update of $V_k(s)$ over all states $s$, and
        \item \textit{lower bound} the expected one state update of $V_k(s)$ over all states $s \not\in S$
        as above $\epsilon > 0$.
    \end{itemize}
    
    Let's start by lower bounding $b(s)$, or equivalently upper bounding $V_k(s)$.
    Note that the maximum value of $V_k(s)$ must be attained by a state $s$ such that the maximum possible reward $r_{\max}(s)$ exceeds $J_k$,
    as can be seen by examining the Poisson equation \eqref{eq:poisson-v}.
    There are only finitely many such states, by \cref{ass:reward}(b), so $V_k(s)$ is upper bounded.
    Note that $J_k$ is finite because $\pi_k$ is stable by our inductive assumption.
    
    Next, let's look at the expected value of $V_k(s)$, after a one-step transition according to policy $\pi_{k+1}$,
    and apply \eqref{eq:poisson-q}, the Poisson equation:
    \begin{align}
        \label{eq:stable-mid}
        E_{s' \sim P_{k+1}(s)}[V_k(s')] = Q_k(s, \pi_{k+1}) + J_k - r(s, \pi_{k+1}).
    \end{align}
    
    Now, we apply \cref{lem:monotonic-q}, which states that when measured according to $\pi_k$'s relative value function $Q_k(s, a)$,
    policy $\pi_{k+1}$'s actions have average relative value at least that of $\pi_k$:
    \begin{align*}
        \forall s, \, Q_k(s, \pi_{k+1}) \ge Q_k(s, \pi_k) = V_k(s).
    \end{align*}
    Note that the \cref{lem:monotonic-q} does not require $\pi_{k+1}$ to be a stable policy, so we may use \cref{lem:monotonic-q}.
    Combining \cref{lem:monotonic-q} with \eqref{eq:stable-mid},
    \begin{align*}
        E_{s' \sim P_{k+1}(s)}[V_k(s')]  \ge V_k(s) + J_k - r(s, \pi_{k+1}).
    \end{align*}
    
    Let $S$ be the set of states $s$ with maximum reward $r_{\max}(s) \ge J_k - 1$.
    Note that $S$ is a finite set, by \cref{ass:reward}(b).

    We then have the following bounds on the one-step update of $V_k(s)$ under policy $\pi_{k+1}$:
    \begin{align*}
        \forall s,\, E_{s' \sim P_{k+1}(s)}[V_k(s')] - V_k(s) &= J_k - r(s, \pi_{k+1}) \ge J_k - c_{\max} \\
        \forall s \not\in S,\, E_{s' \sim P_{k+1}(s)}[V_k(s')] - V_k(s) &= J_k - r(s, \pi_{k+1}) \ge J_k - (J_k - 1) = 1.
    \end{align*}
    
    By \textit{lower bounding} the expected change of $V_k(s)$, we equivalently upper bound the expected change of $b(s) = -V_k(s)$:
    \begin{align*}
        \forall s,\, E_{s' \sim P_{k+1}(s)}[b(s')] - b(s) &\le c_{\max} - J_k \\
        \forall s \not\in S,\, E_{s' \sim P_{k+1}(s)}[b(s')] - b(s) &\le -1.
    \end{align*}
    Setting $C = c_{\max} - J_k$ and $\epsilon = 1$,
    all conditions on $b(s)$ are satisfied,
    and by the Foster-Lyapunov theorem, policy $\pi_{k+1}$ is stable. By induction, all NPG iterates are stable.
\end{proof}
\subsection{Step 1: Bounding change in relative value under NPG update}

Next, we bound the amount by which the relative value function of an NPG iterate $V_{k+1}(s)$ can be lower (i.e. worse)
than $V_k(s)$, the previous policy iterate, relative to $\tau_{k+1}(s)$, the time to hit $\sz$ from state $s$ under policy $\pi_{k+1}$:

\begin{lemma}
    \label{lem:rel-policy}
    For any two NPG policy iterates $\pi_k$ and $\pi_{k+1}$, and any state $s$, we can lower bound $V_{k+1}(s)$:
    \begin{align*}
        V_{k+1}(s) \ge V_k(s) - \tau_{k+1}(s)(J_{k+1} - J_k).
    \end{align*}
\end{lemma}
\begin{proof}
    To compare $V_{k+1}$ and $V_k$, we will examine non-stationary policies which perform
    policy $\pi_{k+1}$ for some number of steps, and then perform $\pi_k$ afterwards.
    Note that such policies have an average reward of $J_k$,
    so their relative values are directly comparable to $V_k(s)$.

    Let $V_k^n(s)$ denote the relative value of the policy which performs $\pi_{k+1}$ for $n$ steps before switching to $\pi_k$.
    Note that $V_k^0(s) = V_k(s)$, and that $V_k^1(s) = Q_k(s, \pi_{k+1})$.

    Applying \cref{lem:monotonic-q}, we know that $V_k^1(s) \ge V_k^0(s)$.
    Furthermore, let us compare $V_k^2(s)$ and $V_k^1(s)$.
    Both policies start by applying $\pi_{k+1}$ for one step,
    accruing the same reward and transitioning to the same distribution over states.
    From that point onward, \cref{lem:monotonic-q} again tells us that $V_k^2(s) \ge V_k^1(s)$.
    In general, $V_k^n(s) \ge V_k^0(s)$ for any number of steps $n$.
    Even if $n$ is chosen to be a stopping time, rather than a constant number of steps,
    $V_k^n(s) \ge V_k^0(s)$.

    Let us consider the specific case where the number of steps $n$ is the time to hit $\sz$ from state $s$
    under the policy $\pi_{k+1}$. This policy has relative value $V_k^{\tau_{k+1}(s)}(s)$.

    Because we define $V_\pi(\sz) = 0$, for any policy $\pi$,
    to compute a policy's relative value, we need only examine its relative value
    over the first $\tau_\pi(s)$ steps.
    In particular,
    \begin{align*}
        V_\pi(s) = \Ep\left[ \sum\nolimits_{i=0}^{\tau_\pi(s)} \left( r(s_\pi(i), a_\pi(i)) - J_\pi \right)\right],
    \end{align*}
    where $s_\pi(i), a_\pi(i)$ are the state and action taken by policy $\pi$ on iteration $i$.

    Applying this formula for $V_k^{\tau_{k+1}(s)}(s)$, we find that
    \begin{align*}
        V_k^{\tau_{k+1}(s)}(s) &= \Ep\left[ \sum\nolimits_{i=0}^{\tau_{k+1}(s)} \left( r(s_{k+1}(i), a_{k+1}(i)) - J_k \right)\right] \\
        &= V_{k+1}(s) + \tau_{k+1}(s) (J_{k+1} - J_k).
    \end{align*}
    Applying the fact that $V_k^{\tau_{k+1}(s)}(s) \ge V_k(s)$,
    we find that
    \begin{align*}
        V_{k+1}(s) + \tau_{k+1}(s) (J_{k+1} - J_k) &= V_k^{\tau_{k+1}(s)}(s) \ge V_k(s) \\
        V_{k+1}(s) &\ge V_k(s) - \tau_{k+1}(s) (J_{k+1} - J_k).
        %\qedhere
    \end{align*}
\end{proof}

\subsection{Step 2: Bounding the time to hit the empty-queue state $\sz$.}

Now, we bound the time $\tau_\pi(s)$ for the system to hit the empty queue state $\sz$,
or more generally the highest-reward state,
starting from any state $s$ and under any policy $\pi$.
Later, we will apply this bound when the initial state $s$ has high reward $r_{\max}(s)$
and for policies $\pi$ with high average reward $J_\pi$.
In proving our bound, we make key use  of \cref{ass:connected}, our assumption of uniform connectedness
of states with high reward $r_{\max}(s)$, e.g. short queues.

\begin{lemma}
    \label{lem:uniform-pr}
    Consider an MDP satisfying \cref{ass:connected,ass:reward}.
    Let $y, z$ be two reward thresholds, $y > z$.
    For all states $s$ with $r_{\max}(s) \ge z$
    and all stable policies $\pi$ such that $J_\pi \ge y$,
    the hitting time is bounded:
    \begin{align*}
        \tau_\pi(s) \le \tau^{bound}_{y,z} := \frac{x_z (c_{\max} - y)}{p_z^2(y-z)} + x_z,
    \end{align*}
    where $x_z$ and $p_z$ are the constants from \cref{ass:connected}.
\end{lemma}
\begin{proof}
    Let $s$ be a generic state such that $r_{\max}(s) \ge z$.
    As $\pi$ is stable, it returns to $s$ infinitely often. Let us consider the renewal cycle with renewal point in state $s$.

    Let $\s_z$ be the set of states $s'$ with $r_{\max}(s') \ge z$.
    
    We will start by upper bounding the fraction of each cycle that is spent in states $s_t \not\in \s_z$. This will be used to construct an upper bound on the hitting time to reach the identified state $\sz$ (e.g. the empty queue).

    Let $T_{in}^s$ denote the expected time per $s$-cycle that the system spends in states $s_t \in \s_z$, and let $T_{out}^s$ denote the expected time per cycle with $s_t \not\in \s_z$.
    We can bound the total expected reward per cycle as
    \begin{align*}
        \Ep[\text{total reward}] \le c_{\max} T^s_{in} + z T^s_{out}
    \end{align*}
    Knowing that the expected mean reward $J_\pi$ is at least $y$, we apply the Renewal-Reward theorem \cite[Theorem 23.4]{harchol_performance_2013}:
    \begin{align}
        y &\le J_\pi = \frac{\Ep[\text{total reward}]}{T_{in}^s + T^s_{out}} \le \frac{c_{\max} T^s_{in} + z T^s_{out}}{T^s_{in} + T^s_{out}} \implies
        %y (T_{in} + T_{out}) &\le c_{\max} T_{in} + z T_{out} \\
        \label{eq:t-in-t-out}
        T^s_{out} \le \frac{c_{\max} - y}{y-z} T^s_{in} %\\
        %\frac{y-z}{c_{\max} - y} &\le \frac{T_{in}}{T_{out}}
    \end{align}

    Next, we'll directly prove an upper bound on $T^s_{in}$, which in turn upper bounds $T^s_{out}$.
    By \cref{ass:connected}, starting on each time step $t$ on which $s_t \in \s_z$, there is at least a probability $p_z$ that in the interval consisting of the next $x_z$ steps,
    the system reaches the renewal state $s$.
    There can be at most $\frac{1}{p_z}$ such intervals in expectation per renewal period, and all time steps on which $s_t \in \s_z$ must be contained in these intervals.
    As a result, $T^s_{in} \le \frac{x_z}{p_z}$.
    Applying \eqref{eq:t-in-t-out}, we find that
    \begin{align*}
        T^s_{out} \le \frac{c_{\max} - y}{y - z} \frac{x_z}{p_z}
    \end{align*}

    Note that this bound on $T^s_{out}$ applies for any state $s \in \s_z$. Note also that $T^s_{out}$ is an upper bound on the expected time between visits to $\s_z$.

    Let $p_{\pi,z}(s)$ be the probability that the system reaches $\sz$ from a generic state $s \in \s_z$ in at most $x_z$ steps.
    Let $\{T^s_{out}\}'$ be the expected time, starting from state $s$, until the system re-enters $\s_z$, conditional on not reaching $\sz$ in the first $x_z$ steps.
    Note that 
    \begin{align*}
        {T^s_{out}}' \le \frac{T^s_{out}}{1-p_{\pi,z}(s)},
    \end{align*}
    because the probability of the event of  not reaching $\sz$ in at most $x_z$ steps
    is $1-p_{\pi,z}(s)$.
    We can now start to bound $\tau_\pi(s)$ as follows, based on whether or not we hit $\sz$ on the current visit to $\s_z$:
    \begin{align*}
        \tau_\pi(s) \le p_{\pi,z}(s) x_z + (1-p_{\pi,z}(s)) ({T^s_{out}}' + \Ep[\tau_\pi(s')]),
    \end{align*}
    where $s'$ is a random variable denoting the state at which we re-enter $\s_z$. Thus,
    \begin{align*}
        \tau_\pi(s) &\le p_{\pi,z}(s) x_z + (1-p_{\pi,z}(s)) \left(\frac{T^s_{out}}{1-p_{\pi,z}(s)} + \Ep[\tau_\pi(s')] \right) \\
            %&= p_{\pi,z}(s) x_z + t_{\pi,z}(s) + (1-p_{\pi,z}(s)) \tau_\pi(s') \\
            &= p_{\pi,z}(s) x_z + T^s_{out} + (1-p_{\pi,z}(s)) \Ep[\tau_\pi(s')] \\
            &= x_z + T^s_{out} + (1-p_{\pi,z}(s))(\Ep[\tau_\pi(s')] - x_z)
    \end{align*}
    
    In particular, letting $\tau_\pi^{\max}$ be the maximum over $s \in \s_z$ of $\tau_\pi(s)$,
    and letting $s^*$ be the state in which that maximum occurs,
    we have
    \begin{align*}
        \tau_\pi^{\max} &\le x_z + T^{s^*}_{out} + (1-p_{\pi,z}(s^*))(\tau_\pi^{\max} - x_z) \\
        %p_{\pi,z}(s_{\tau}^*) (\tau_\pi^{\max} - x_z) &\le t_{\pi,z}(s_{\tau}^*) \\
        \tau_\pi^{\max} &\le x_z + \frac{T^{s^*}_{out}}{p_{\pi,z}(s^*)}
        %\le x_z + \frac{t_{\pi,z}^{\max}}{p_z}
        \le \frac{x_z (c_{\max} - y)}{p_z^2(y-z)} + x_z.
        %\qedhere
    \end{align*}
\end{proof}

\subsection{Step 3: Bounding relative value $V_k(s)$ for all states $s$}

We are now ready to bound the relative value function $V_k$ for all iterates $\pi_k$ of the Natural Policy Gradient algorithm,
and for all states $s$. \cref{lem:uniform-pr} covered high-reward states, so this lemma focuses on low-reward states,
building off of \cref{lem:rel-policy} to do so.
\begin{lemma}
    \label{lem:bound-vk}
    For any reward threshold $z < J_0$,
    and for each iterate $\pi_k$ of the NPG algorithm,
    the relative value function is lower and upper bounded as follows:
    \begin{align*}
        \forall s,k, \quad &V_k(s) \ge\min(V_0(s) \frac{J_* - z}{J_0 - z}, V_0(s)) - \tau^{bound}_{J_0,z} \frac{(J_* - J_0)(c_{\max} - z)(J_* - z)}{(J_0 - z)^2} \\
        \forall s,k, \quad &V_k(s) \le \tau^{bound}_{J_0,z} (c_{\max} - J_0),
    \end{align*}
    where $\tau^{bound}_{y,z}$ is the hitting-time bound from \cref{lem:uniform-pr}.
    Note that these bounds do not depend on the iteration $k$,
    and they only depend on the state $s$ via the initial policy's relative value $V_0(s)$.
\end{lemma}
\begin{proof}

To lower bound $V_k(s)$, let us start by using the following formula for $V_k(s)$:
\begin{align*}
    V_k(s) = - J_k \tau_k(s) + \rbar_k(s) \tau_k(s)
\end{align*}
where $\rbar_k(s)$ is the average reward that policy $\pi_k$ accrues over the interval until it first enters state $\sz$.

Let us subdivide the time $\tau_k(s)$ into two periods:
Let $\tau_k^{out}$ be the time until the policy $\pi_k$ first enters $\s_z$,
and let $\tau_k^{in}$ be the time from then until the policy first enters state $\sz$.
We define $\rbar_k^{out}$ and $\rbar_k^{in}$ similarly.

Similarly, let us define $V_k^{out}(s)$ and $V_k^{in}(s)$:
\begin{align*}
    V_k^{out}(s) &= - J_k \tau_k^{out}(s) + \rbar_k^{out}(s) \tau_k^{out}(s), 
    V_k^{in}(s) = - J_k \tau_k^{in}(s) + \rbar_k^{in}(s) \tau_k^{in}(s)
\end{align*}
Note that $V_k(s) = V_k^{out}(s) + V_k^{in}(s)$

We can upper bound each of these quantities. Note that $\rbar_k^{out}(s) \le z$, and $\rbar_k^{in}(s) \le c_{\max}$.
Note that $\tau_k^{in}(s) \le \tau^{bound}_{J_0,z}$,
by \cref{lem:uniform-pr}.
Together, we can upper bound $V_k(s)$ relative to $\tau_k(s)$, or equivalently upper bound $\tau_k(s)$ relative to $V_k(s)$:
\begin{align*}
    V_k(s) &\le -J_k \tau_k(s) + z \tau_k(s) + (c_{\max} - z) \tau^{bound}_{J_0,z} \\
    %V_k(s) - (c_{\max} - z) \tau^{bound}_{J_0,z} &\le \tau_k(s) (-J_k + z) \\
    \frac{V_k(s) - (c_{\max} - z) \tau^{bound}_{J_0,z}}{-J_k + z} &\ge \tau_k(s)
\end{align*}
Recall that by \cref{lem:monotonic-j}, $J_\pi \ge J_0 > z$, so $-J_k + z < 0$, so the direction of the inequality flips.
Here we use \cref{lem:stable}, which states that all NPG iterates are stable.

The key fact relating $V_k(s)$ and $V_{k+1}(s)$ is \cref{lem:rel-policy}:
$V_{k+1}(s) \ge V_k(s) - \tau_{k+1}(s)(J_{k+1} - J_k)$.

Combining our bounds, we find that
\begin{align*}
    V_{k+1}(s) &\ge V_k(s) - \frac{V_{k+1}(s) - (c_{\max} - z) \tau^{bound}_{J_0,z}}{-J_{k+1} + z}(J_{k+1} - J_k) \\
    %V_{k+1}(s) + \frac{V_{k+1}(s)}{-J_{k+1} + z}(J_{k+1} - J_k) &\ge V_k(s) - \frac{(c_{\max} - z) \tau^{bound}_{J_0,z}}{-J_{k+1} + z}(J_{k+1} - J_k) \\
    %V_{k+1}(s) \frac{-J_{k+1} + z + J_{k+1} - J_k}{-J_{k+1} + z} &\ge V_k(s) + (J_{k+1} - J_k)\frac{(c_{\max} - z) \tau^{bound}_{J_0,z}}{J_{k+1} - z} \\
    %V_{k+1}(s) \frac{J_k - z}{J_{k+1} - z} &\ge V_k(s) + (J_{k+1} - J_k)\frac{(c_{\max} - z) \tau^{bound}_{J_0,z}}{J_{k+1} - z} \\
    V_{k+1}(s) &\ge V_k(s) \frac{J_{k+1} - z}{J_k - z} - \tau^{bound}_{J_0,z}(J_{k+1} - J_k)\frac{(c_{\max} - z)}{J_k - z}
\end{align*}

Now, we apply this bound telescopically, for all $\pi_i \in [0, k]$. By doing so, we find that
\begin{align*}
    V_k(s) \ge V_0(s) \frac{J_k - z}{J_0 - z} - \tau^{bound}_{J_0,z} \sum_{i=0}^{k-1} (J_{i+1} - J_i) \frac{c_{\max} - z}{J_i - z} \frac{J_k - z}{J_{i+1} - z}
\end{align*}
Applying the monotonicity bound $J_0 \le J_k$ (\cref{lem:monotonic-j}) and the optimality bound $J_k \le J_*$,
\begin{align*}
    %V_k(s) &\ge V_0(s) \frac{J_k - z}{J_0 - z} - \tau^{bound}_{J_0,z} \sum_{i=0}^{k-1} (J_{i+1} - J_i) \frac{c_{\max} - z}{J_0 - z} \frac{J_* - z}{J_0 - z} \\
    %V_k(s) &\ge V_0(s) \frac{J_k - z}{J_0 - z} - \tau^{bound}_{J_0,z} (J_k - J_0) \frac{c_{\max} - z}{J_0 - z} \frac{J_* - z}{J_0 - z} \\
    V_k(s) &\ge \min(V_0(s) \frac{J_* - z}{J_0 - z}, V_0(s)) - \tau^{bound}_{J_0,z} \frac{(J_* - J_0)(c_{\max} - z)(J_* - z)}{(J_0 - z)^2}
\end{align*}
%\isaac{Is there a better way to handle the possibility that $V_0(s) > 0$ than with the min?}

That completes the proof of the lower bound.

To upper bound $V_k(s)$, let's again focus on $V_k^{out}(s)$ and $V_k^{in}(s)$.
\begin{align*}
    V_k^{out}(s) &= - J_k \tau_k^{out}(s) + \rbar_k^{out}(s) \tau_k^{out}(s)
        \le - J_k \tau_k^{out}(s) + z \tau_k^{out}(s) \\
        &= (- J_k + z) \tau_k^{out}(s) \le 0 \\
    V_k^{in}(s) &= - J_k \tau_k^{in}(s) + \rbar_k^{in}(s) \tau_k^{in}(s)
        \le (c_{\max} - J_k) \tau_k^{in}(s)
        %&\le (c_{\max} - J_k) \tau^bound_{J_0,z} \\
        \le (c_{\max} - J_0) \tau^{bound}_{J_0,z} \\
    V_k(s) &= V_k^{out}(s) + V_k^{in}(s) \le (c_{\max} - J_0) \tau^{bound}_{J_0,z}
    %\qedhere
\end{align*}
\end{proof}

Now, we combine \cref{lem:bound-vk} with an assumption on the initial policy and our mild structural assumption on the MDP, \cref{ass:reward},
to give a state-dependent bound on $Q_k(s, a)$:
\begin{lemma}
    \label{lem:bound-vk-state}
    Given an MDP satisfying \cref{ass:reward},
    and given an initial policy $\pi_0$ such that
    there exist constants $c_0 > 0, c_1 \ge 0$ such that
    \begin{align*}
        V_0(s) \ge -c_0 \rhat_{\max}(s)^2 - c_1,
    \end{align*}
    there exists a uniform bound $M_s$:
    \begin{align*}
        M_s := c_2 \vert \rhat_{\max}(s)\vert ^2 + c_3 \vert \rhat_{\max}(s)\vert  + c_4,
    \end{align*}
    for constants $c_2, c_3, c_4 \ge 0$ depending on $c_0, c_1$, and the MDP parameters,
    such that for any NPG iterate $\pi_k$, and any pair of actions $a, a'$,
        \begin{align*}
            Q_k(s, a) - Q_k(s, a') \le M_s.
        \end{align*}
\end{lemma}
\begin{proof}
    Deferred to \cref{app:bound-vk-state}.
\end{proof}

\subsection{Step 4: Proof of general result}
\label{sec:main-proof}

With our bound \cref{lem:bound-vk-state} on the relative value function $V_k(s)$ of the iterates of the NPG algorithm,
we are now ready to prove our main result on the convergence of the NPG algorithm, \cref{thm:general-main}.

\begin{reptheorem}{thm:general-main}
    For any average-reward MDP satisfying \cref{ass:reward,ass:connected},
    given an initial policy $\pi_0$ such that
    there exist constants $c_0 > 0, c_1 \ge 0$ such that
    \begin{align}
        \label{eq:v-ass}
        V_0(s) \ge -c_0 \rhat_{\max}(s)^2 - c_1,
    \end{align}
    the NPG algorithm with learning rate parameterization    
    $\beta_s$ given in \eqref{eq:beta-s-main}
    achieves the convergence rate $J_* - J_T \le \frac{c_*}{\sqrt{T}}$,
    where $c_*$ is
    a constant depending on the MDP parameters and on $c_0$ and $c_1$.
\end{reptheorem}

\begin{proof}
    As outlined in \cref{sec:proof-sketch},  we will think of the MDP optimization problem as many instances
    of the expert advice problem with reward $Q_k(s, a)$.
    Note that the NPG algorithm, \cref{alg:npg}, is exactly identical to the weighted majority algorithm,
    executing a parallel instance of the weighted majority algorithm in every state $s$.

    Moreover, the MDP objective of maximizing the average reward $J_\pi$
    is closely related to the objective of maximizing total reward in the expert advice problem.
    Recall the Performance Difference Lemma, \cref{lem:perf-diff},
    which states that for any pair of policies $\pi, \pi'$:
    \begin{align*}
        J_{\pi} - J_{\pi'}  = E_{s \sim d_\pi} [Q_{\pi'}(s, \pi) - V_{\pi'}(s)].
    \end{align*}
    Let us apply this lemma for the iterates $\pi_k$ of our algorithm,
    in comparison to the optimal policy $\pi_*$, summing over all iterates $k \in [1, T]$:
    \begin{align*}
        \sum\nolimits_{k=1}^T J_* - J_k &= E_{s \sim \pi_*} \left[\sum\nolimits_{k=1}^T Q_k(s, \pi_*) - V_k(s)\right]
    \end{align*}

    Note that for any specific state $s$, $\sum_{k=1}^T Q_k(s, \pi_*) - V_k(s)$
    is exactly the difference in total reward between a specific fixed policy $\pi_*(a \mid s)$
    and the weighted majority policy $\pi_k(a \mid s)$
    of the expert advice problem 
    with reward function at time step $k$ of $Q_k(s, a)$.
    
    Thus, we can apply \cref{lem:weighted-majority}, a known regret bound on the performance of the weighted majority algorithm
    for the expert advice problem, to bound the convergence rate of the NPG algorithm.
    To do so, we use \cref{lem:bound-vk-state}, which states that for all NPG intermediate policies $\pi_k$ and actions $a, a'$,
    \begin{align*}
        Q_k(s, a) - Q_k(s, a') &\le M_s \\
        \text{where } M_s &:= c_2 \vert \rhat_{\max}(s)\vert ^2 + c_3 \vert \rhat_{\max}(s)\vert  + c_4,
    \end{align*}
    for some positive constants $c_2, c_3, c_4$ given in the proof of \cref{lem:bound-vk-state}.
    In particular, $M_s$ can be computed ahead of time,
    given only the structure of the MDP and the initial policy $\pi_0$.
    
    We select the learning rate $\beta_s$ for state $s$ as given in \cref{lem:weighted-majority}:
    \begin{align}
        \label{eq:beta-s-main}
        \beta_s := g \left(\sqrt{\frac{\ln \vert A_s\vert }{T M_s}} \right),
        \text{where } g(z) = 1 + 2z + z^2/\ln 2.
    \end{align} 
    Thus, we may apply \cref{lem:weighted-majority},
    and thereby obtain the following regret guarantee for the NPG algorithm:
    \begin{align*}
        \sum\nolimits_{k=1}^T Q_k(s, \pi_*) - V_k(s) &\le \sqrt{T M_s \ln \vert A_s\vert } + \log_2(\vert A_s\vert ) \\
        &=
        %\sqrt{T M_s \ln \vert A\vert } + \frac{\log_2\vert A\vert }{2} 
        %=
        \sqrt{T \ln \vert A_s\vert } \sqrt{c_2 \vert \rhat_{\max}(s)^2\vert  + c_3 \vert \rhat_{\max}(s)\vert  + c_4} + \log_2(\vert A_s\vert )/2  \\
        %&\le \sqrt{T \ln \vert A\vert } (\sqrt{c_5} \vert\rhat_{\max}(s)\vert) + \sqrt{c_6} (\rhat_{\max}(s) + 1/4) + \sqrt{c_7}) + \frac{\log_2 \vert A\vert }{2} \\
        &\le \sqrt{T \ln A_{\max} } (c_5 \vert \rhat_{\max}(s)\vert  + c_6) + \log_2(A_{\max} )/2 \\
        \text{where } c_5 &:= \sqrt{c_2} + \sqrt{c_3}, \text{ and } c_6 := \sqrt{c_3}/4 + \sqrt{c_4}.
    \end{align*}

    Now, let's bound the difference in average reward between the iterates and the optimal policy:
    \begin{align}
        \label{eq:using-main-assumption}
        \sum\nolimits_{k=1}^T J_* - J_k
        %&= E_{s \sim \pi_*} \left[\sum_{k=1}^T Q_k(s, \pi_*) - V_k(s)\right]
        \le \sqrt{T \ln A_{\max} } (c_5 E_{s \sim \pi_*}[\vert \rhat_{\max}(s)\vert ] + c_6) + \log_2(A_{\max})/2
    \end{align}
    Here, we see the importance of our quadratic assumption on $\pi_0$, \eqref{eq:v-ass}.
    We know that $J_* = E_{s \sim \pi_*}[r(s, \pi_*(s))]$ is finite and small.
    This is essentially the only property we know about the optimal policy $\pi_*$.
    Because of our quadratic assumption, we have shown that $J_k$ depends crucially on $E_{s \sim \pi_*}[\vert \rhat_{\max}(s)\vert ]$.
    If $\pi_0$ had a faster-growing relative value function, we would only be able to relate $J_k$ to
    $E_{s \sim \pi_*}[\vert \rhat_{\max}(s)\vert ^\alpha]$ for some $\alpha > 1$, which we would not be able to bound.

    Now that we have shown that, under our assumption, the NPG algorithm's regret depends on $E_{s \sim \pi_*}[\vert \rhat_{\max}(s)\vert ]$,
    note that 
        $\vert \rhat_{\max}(s)\vert  \le \vert \rhat(s, a)\vert  = c_{\max}-r(s, a)$ for any action $a$.
    As a result,
    \begin{align*}
            \sum\nolimits_{k=1}^T J_* - J_k 
            &\le \sqrt{T \ln A_{\max} } (c_5 E_{s \sim \pi_*}[c_{\max}-r(s, \pi_*)] + c_6) + \log_2(A_{\max} )/2  \\
            &= \sqrt{T \ln A_{\max} } (c_5 (c_{\max}-J_*) + c_6) + \log_2(A_{\max} )/2 
    \end{align*}
    Thus, because $T \ge 1$,
    we have
    \begin{align*}
        \sum\nolimits_{k=1}^T J_* - J_k \le c_* \sqrt{T}, \text{ where } c_* := \sqrt{\ln A_{\max} } (c_5 (c_{\max} - J_*) + c_6) + \log_2( A )/2.
    \end{align*}
    Finally, by \cref{lem:monotonic-j}, we know that $J_T \ge J_k$ for all iterates $k$.
    As a result, 
    \begin{align*}
        J_* - J_T &\le \frac{c_*}{\sqrt{T}}.
        %\qedhere
    \end{align*}
\end{proof}

\subsection{Bounding Relative Value $V_{MW}(s)$ for MaxWeight}
\label{sec:bound-mw}

First, we show how to bound $V_\pi(s)$ using a Lyanpunov function argument.

\begin{lemma}\label{lem:lyapunov}
    Under a given policy $\pi$,
    suppose that there exists a function $f(s) \ge 0$ and constants $c_1 > 0, c_2 \ge 0$ such that
    \begin{align*}
        \forall s, \quad E_{\pi}[f(s_{t+1}) - f(s_t) \mid s_t = s] \le c_1 r(s, \pi) + c_2.
    \end{align*}
    Then there exist explicit constants $c_3, c_4 \ge 0$ such that $V_\pi(s) \ge -c_3 f(s) - c_4$.
\end{lemma}
\begin{proof}
    We will start by rescaling the function $f(s)$ to more closely relate it to the value function.
    Let $\widetilde{f}(s) = \frac{2}{c_1} f(s)$.
    Then we have
    \begin{align*}
        E_{\pi}[\widetilde{f}(s_{t+1}) - \widetilde{f}(s_t) \mid s_t = s] \le 2 r(s, \pi) + 2 c_2/c_1
    \end{align*}

    Next, let $c_5$ be the reward threshold $c_5 = -\frac{2 c_2}{c_1} - J_\pi$.
    For states $s$ for which $r_{\max}(s) \le c_5$, we have
    \begin{align*}
        E_{\pi}[\widetilde{f}(s_{t+1}) - \widetilde{f}(s_t) \mid s_t = s] \le r(s, \pi) - J_\pi
    \end{align*}

    Next, letting $s_0 = s$, let $\tau^{c_5}$ be the first positive time when the system enters $\s_{c_5}$,
    the set of states for which $r_{\max}(s) \ge c_5$.
    Let $V^{c_5}(s)$ be the expected relative value after this point in time:
    \begin{align*}
        \sum\nolimits_{t=0}^{\tau^{c_5}} E_{\pi}[\widetilde{f}(s_{t+1}) - \widetilde{f}(s_t)] &\le \sum\nolimits_{t=0}^{\tau^{c_5}} r(s_t) - J_\pi \\
        \widetilde{f}(s_{\tau^{c_5}}) - \widetilde{f}(s) &\le V_\pi(s) - V^{c_5}(s) \\
        V_\pi(s) &\ge V^{c_5}(s) + \widetilde{f}(s_{\tau^{c_5}}) - \widetilde{f}(s).
    \end{align*}
    Because $f(s) \ge 0$, we know that $\widetilde{f}(s) \ge 0$.
    Because $\s_{c_5}$ is a finite set, $V^{c_5}$ is bounded below:
    \begin{align*}
        V^{c_5}(s) & \ge V_{\min} := \min_{s \in \s_{c_5}} V(s) \\
        V_\pi(s) &\ge V_{\min} - \widetilde{f}(s) = - \frac{2}{c_1} f(s) + V_{\min}.
    \end{align*}
    Setting $c_3 = \frac{2}{c_1}$ and $c_4 = (-V_{\min})^+$ completes the proof.

\end{proof}

Now, we use the framework established in \cref{lem:lyapunov} to bound the relative value function for the MaxWeight policy.

The stability region of a queueing system is defined to be the set of average arrival rates $\lambda_i$ for which there exists a scheduling policy
such that the system is stable, i.e. the system state converges to a stationary distribution.
The GSSE stability region is the convex hull of the mean service rates $\mu_i^j$.
The optimal stability region is achieved by the MaxWeight policy, defined in \eqref{eq:maxweight-def}.

The GSSE stability region is defined by the following constraints:
For an arrival rate vector $\lambda$ to be stable,
there must exist a nonnegative service vector $\gamma \ge 0$ and a slack margin $\epsilon > 0$ such that
\begin{align}
    \sum\nolimits_j \gamma_j = 1 \quad \text{ and } \quad
    \label{eq:convex}
    \forall i, (1 + \epsilon) \lambda_i \le \sum\nolimits_j \gamma_j \mu_i^j.
\end{align}

We can bound MaxWeight's relative value $V_{MW}(q)$ if we can prove a Lyapunov function result of the form given by \cref{lem:lyapunov}.
We prove such a result using the Lyapunov function $f(q) = \sum_i q^2_i$,
which implies that $V_{MW}(q)$ is $O(q^2)$.

\begin{lemma}
    \label{lem:bound-mw}
    For any GSSE queueing system,
    \begin{align*}
        E_{MW}[f(q(t+1)) - f(q(t)) \mid q(t) = q ] \le c_1 r(q(t)) + c_2,
    \end{align*}
    where $MW$ denotes the MaxWeight policy,
    where $f(q) := \sum_i q^2_i$,
    and where $c_1$ and $c_2$ are:
    \begin{align*}
        c_1 &= 2 \epsilon \min_i \lambda_i, \qquad c_2 = \ell^2 n.
    \end{align*}
\end{lemma}
\begin{proof}
This result is proven in \cite[Theorem 4.2.4]{srikant_communication_2013}, for the classic $n \times n$ switch model.
However, the proof applies unchanged to any GSSE system. For completeness, we reprove the result in the GSSE setting in \cref{app:bound-mw}.    
\end{proof}

\subsection{Rapid convergence for NPG on queueing MDPs}
\label{sec:queueing-main}
\begin{reptheorem}{thm:queueing-main}
    For any GSSE queueing system,
    with the objective of minimizing mean queue length
    starting with the initial policy $\pi_0 = $MaxWeight,
    the NPG algorithm with learning rate parameterization $\beta_s$ given in
    \cref{thm:general-main} achieves the convergence rate $E[ \vert q_{\pi_T} \vert_1] \le E[\vert q_*\vert_1] + \frac{c_*}{\sqrt{T}}$
    for a constant $c_*$ given in the proof of \cref{thm:general-main}. \\
    This result generalizes to the $\alpha$th moment of queue length.
\end{reptheorem}
\begin{proof}
To prove this result, we will apply \cref{thm:general-main}, with reward equal to negative total queue length $-\vert q \vert_1$.
To do so, we will verify \cref{ass:reward,ass:connected}
and prove the MaxWeight achieves the required bound \eqref{eq:v-ass} on the relative value function.

Let us start by verifying \cref{ass:reward}.
\begin{itemize}
    \item\cref{ass:reward}(a): The maximum possible reward is 0, achieved by the all-zeros state $q_i=0$.
    \item\cref{ass:reward}(b): The set of states with total reward at least $z$ is the set of states with total queue length at most $\vert z\vert $. As a generous bound, there are at most $n^{\vert z\vert }$ such states, recalling that $n$ is the number of job classes.
    \item\cref{ass:reward}(c): In each state, the reward under all actions is equal.
    \item\cref{ass:reward}(d): If it is possible to transition from state $s$ and $s'$, then $s$ and $s'$ have similar numbers of jobs, differing by at most $\ell n$,
    where $n$ is the number of job classes and $\ell$ is the maximum arrivals or departures per class in a step.
As a result, the maximum reward in those states also differs by at most $\ell n$.
\end{itemize}

Now, let's turn to \cref{ass:connected}, the assumption that the set of states with at most $\vert z\vert $ jobs present is uniformly connected under an arbitrary policy $\pi$.

In particular, we will show that there is a nonzero probability that, for any pair of states $q, q'$ with at most $\vert z\vert $ jobs present,
the system moves from $q$ to $q'$ in at most $2\vert z\vert $ time steps.
More specifically, there is a nonzero probability that the system moves from $q$ to $\sz$ in at most $\vert z\vert $ time steps,
and then moves from $\sz$ to $q'$ in at most $\vert z\vert $ time steps.

To see why, first recall from \cref{sec:def-gsse} that our action space consists solely of non-idling actions:
If the system state is not the all-zero state,
the only service options in the action space are service options which have a nonzero probability of completing a job.
Recall also that on each time step, there's a nonzero probability that no new job arrives.
Thus, on each of the first $\vert z\vert $ time steps, if the system is not yet empty
there is a nonzero probability that at least one job completes and no new job arrives.
At some point during those steps, we will reach the all-zeros state.

Recall also that we require that for any service option $j$, for each class $i$ there is a nonzero chance that more jobs arrive than complete,
and a nonzero chance that the same number of jobs arrive as complete.
Thus, over the next $\vert z\vert $ time steps, regardless of the policy selected, there is a nonzero chance that the number of jobs present of class $i$
rises from $0$ to $q_i'$, and then stays at $q_i'$, allowing the system to reach the state $q$.
This verifies \cref{ass:connected}.

Finally, let us verify that the MaxWeight initial policy satisfies the desired relationship,
\begin{align*}
    V_{MW}(s) \ge -c_0 \rhat_{\max}(s)^2 - c_1,
\end{align*}
for some constants $c_0$ and $c_1$.

To do so, we simply apply the Lyapunov function argument from \cref{lem:lyapunov},
using the Lyapunov function $f(q) = \sum_i q_i^2$, which we verified is a valid Lyapunov function in \cref{lem:bound-mw}.

Thus, \cref{thm:general-main} applies to any GSSE queueing MDP, given the MaxWeight initial policy.

This result can be generalized to other reward functions, including a weighted queue length reward,
as well as the $\alpha$-moment reward function $-(\sum_i q_i)^\alpha$, for any $\alpha \ge 1$.
For these generalizations of the reward function, we can use weighted-MaxWeight and $\alpha$-MaxWeight as the initial policies. These policies replace the $q_i$ term in the MaxWeight definition \eqref{eq:maxweight-def} with the weighted queue length or $\alpha$th power of queue length.
The proof that weighted-MaxWeight has the desired relative value property \eqref{eq:v-ass} is essentially immediate, and for $\alpha$-MaxWeight one can use the Lyapunov function $f_{\alpha}(q) = \sum_i q_i^{\alpha+1}$.
\cref{ass:connected,ass:reward} are straightforward to verify for the weighted queue length setting, and 
\cref{ass:reward}(a-c) and \cref{ass:connected} are straightforward to verify for the $\alpha$th moment reward function.
\cref{ass:reward}(d) is slightly more involved, but we verify that it holds in \cref{app:verify}.
\end{proof}

\section{Infinite State, Discounted Reward}
\label{sec:discounted}

In this section, we discuss the discounted setting: Rather than seeking to maximize the average reward, we seek to maximize the $\gamma$-discounted reward, for some $\gamma < 1$, in expectation over some starting distribution $\eta$. In this section, we demonstrate that existing finite-state results generalize to the infinite-state setting.
In doing so, we remove \cref{ass:reward,ass:connected}: They are not needed in the discounted setting, and we will specify the assumptions that are needed.
Moreover, we remove any assumptions on the initial policy $\pi_0$: Any initialization suffices.
We still work in the infinite-state-space, finite-action-space MDP setting specified in \cref{sec:mdp-model}.

The $\gamma$-discounted reward $b^\pi_\gamma(\eta)$ under policy $\pi$ and initial state distribution $\eta$
is defined as follows:
\begin{align}
    \label{eq:disc-reward-def}
    b^\pi_\gamma(\eta) = \sum_{t=0}^\infty \gamma^t \Ep[r(s_t, \pi)],
\end{align}
where $s_t$ is a random variable denoting the state after $t$ steps under policy $\pi$ and initial state distribution $\eta$.
The objective in the discounted setting is to find the policy $\pi$ which maximizes the discounted reward $b^\pi_\gamma(\eta)$
for a fixed initial state distribution $\eta$.

Note that $b^\pi_\gamma(\eta)$ may not converge for some policies $\pi$ or some initial distributions $\eta$.
On the other hand, $b^\pi_\gamma(\eta)$ may converge even for unstable policies $\pi$: For instance, if $0 \le r(s, a) \le 1$ for all states $s$ and actions $a$,
then $0 \le b^\pi_\gamma(\eta) \le \frac{1}{1-\gamma}$ for all policies $\pi$ and all initial state distributions $\eta$.

In this section, we start with the following assumption on $b^\pi_\gamma$:
\begin{assumption}
    \label{ass:bounded-disc-reward}
     We assume that for each initial state $s$, there exist upper and lower bounds $b_\gamma^{\min}(s)$ and $b_\gamma^{\max}(s)$ on the discounted reward of all policies $\pi$:
    \begin{align*}
        \forall \pi, b_\gamma^{\min}(s) \le b^\pi_\gamma(s) \le b_\gamma^{\max}(s)
    \end{align*}
\end{assumption}

We also define the discounted state visitation distribution $d^{\pi}_\eta$:
\begin{align*}
    d^{\pi}_\eta(s) := (1-\gamma) \sum_{t=0}^\infty \gamma^t P(s_t = s \mid \eta, \pi).
\end{align*}
Note that $d^{\pi}_\eta$ is well-defined for all policies $\pi$, including unstable policies.

We also make a second, slightly stronger assumption,
namely that the expected discounted reward over any policy's discounted state visitation distribution is also finite:
\begin{assumption}
    \label{ass:disc-visit-reward}
     We assume that for each initial state $s$, there exist upper and lower bounds $c_\gamma^{\min}(s)$ and $c_\gamma^{\max}(s)$
     on the expected discounted reward of any policy $\pi'$ over the discounted state visitation distribution of any policy $\pi$, starting in state $s$:
    \begin{align*}
        \forall \pi, \pi',  c_\gamma^{\min}(s) \le \Ep[b^{\pi'}_\gamma(d_s^\pi)] \le c_\gamma^{\max}(s)
    \end{align*}
\end{assumption}

For example, both assumptions are satisfied by the GSSE MDPs defined in \cref{sec:def-gsse}.
Because at most $\ell n$ jobs can arrive per time step, $r(s_t, \pi)$ is bounded by a linear function of $t$ for any policy $\pi$.
This linear growth is overwhelmed by the exponential decay of $\gamma^t$,
and thus \eqref{eq:disc-reward-def} converges for any policy $\pi$, including unstable policies.
This confirms \cref{ass:bounded-disc-reward}.
A similar argument also confirms \cref{ass:disc-visit-reward} for GSSE MDPs, again using the linear bound on $r(s_t, \pi)$ for all $\pi$.

With these assumptions, we demonstrate that \cite[Theorem 16]{agarwal_theory_2021}, a standard proof of convergence to optimality for NPG in the finite-state discounted-reward setting,
generalizes to the infinite-state discounted-reward setting.
\begin{theorem}(Generalization of \cite[Theorem 16]{agarwal_theory_2021})
    For any infinite-state finite-action discounted-reward MDP for which \cref{ass:bounded-disc-reward,ass:disc-visit-reward} hold,
    the NPG algorithm achieves convergence rate $O(1/T)$ to the global optimum:
    \begin{align*}
        b^{\pi_*}_\gamma(\eta) - b^{\pi_T}_\gamma(\eta) \le \frac{\ln A_{\max}}{\beta T} + \frac{\Ep[c^{\max}_\gamma(\eta) - c^{\min}_\gamma(\eta)]}{(1-\gamma)T}
    \end{align*}
\end{theorem}
\begin{proof}
We start by applying performance difference lemma to policies $\pi_*$, the optimal policy, and $\pi_k$, the $k$th NPG iterate.
In the discounted reward setting, the performance difference lemma states:
\begin{align}
    \label{eq:disc-perf-diff}
    b^{\pi_*}_\gamma(\eta) - b^{\pi_k}_\gamma(\eta) = \frac{1}{1-\gamma} \Ep_{s\sim d^{\pi_*}_\eta}[Q_{\pi_k}(s, \pi^*) - V_{\pi^*}(s)].
\end{align}
Note that $b^{\pi_*}_\gamma(\eta)$ and $b^{\pi_k}_\gamma(\eta)$ are well-defined by \cref{ass:bounded-disc-reward},
and note that the proof of the performance difference lemma \cite[Lemma 2]{agarwal_theory_2021}
holds in the infinite-state-space setting, even with unstable policies,
because $d^{\pi_*}_\eta$ is well-defined whether or not $\pi_*$ is stable.

We now use the definition of the NPG algorithm in the discounted setting \cite[Lemma 15]{agarwal_theory_2021}:
    \begin{align*}
    \pi_{k+1}(a \mid s) &= \pi_k(a \mid s) \beta^{(Q_k(s, a)-V_k(s))/(1-\gamma)} / Z_{s,k}, \\
    \text{where } Z_{s,k} &= \sum_{a'} \pi_k(a' \mid s) \beta^{(Q_k(s, a')-V_k(s))/(1-\gamma)}.
\end{align*}

Solving for $Q_k(s, a)$, we find that
\begin{align*}
    Q_k(s, a)-V_k(s) = \frac{1-\gamma}{\ln \beta}\ln \frac{\pi_{k+1}(a \mid s)Z_{s,k}}{\pi_k(a \mid s)}
\end{align*}
Substituting into \eqref{eq:disc-perf-diff}, we have
\begin{align*}
    b^{\pi_*}_\gamma(\eta) - b^{\pi_k}_\gamma(\eta) = \frac{1}{\ln \beta} \Ep_{s\sim d^{\pi_*}_\eta}[\sum_a \pi_{k+1}(a \mid s) \ln \frac{\pi_{k+1}(a \mid s) Z_{s, k}}{\pi_k(a \mid s)}]
\end{align*}
Rewriting in terms of the KL divergence, we have
\begin{align}
    \label{eq:disc-kl}
    b^{\pi_*}_\gamma(\eta) - b^{\pi_k}_\gamma(\eta) =  \frac{1}{\ln \beta} \Ep_{s\sim d^{\pi_*}_\eta}[
    KL(\pi_*(s \vert \cdot) \| \pi_k(s \vert \cdot)) - KL(\pi_*(s \vert \cdot) \| \pi_{k+1}(s \vert \cdot))
    + \ln Z_{s, k}]
\end{align}

Now, we apply \cite[Lemma 17]{agarwal_theory_2021}, which relates $Z_{s, k}$ to the change in discounted reward under the NPG policy.
The proof of \cite[Lemma 17]{agarwal_theory_2021} generalizes to the infinite-state setting without alteration.
It states, using our notation, that for all distributions $\mu$:
\begin{align}
    \label{eq:disc-improvement}
    b^{\pi_{k+1}}_\gamma(\mu) - b^{\pi_k}_\gamma(\mu) \ge \frac{1-\gamma}{\ln \beta} E_{s \sim \mu} [\ln Z_{s, k}] \ge 0
\end{align}

Applying \eqref{eq:disc-improvement} and \eqref{eq:disc-kl}, and specifically the fact that $b^{\pi_{k+1}}_\gamma(\eta) \ge b^{\pi_k}_\gamma(\eta)$, by \eqref{eq:disc-improvement}, we find:
\begin{align*}
    &b^{\pi_*}_\gamma(\eta) - b^{\pi_{T-1}}_\gamma(\eta) \le \frac{1}{T} \sum_{t=0}^{T-1} (b^{\pi_*}_\gamma(\eta) - b^{\pi_k}_\gamma(\eta)) \\
    &\le \frac{1}{T \ln \beta} \sum_{t=0}^{T-1} \Ep_{s\sim d^{\pi_*}_\eta}[KL(\pi_*(s \vert \cdot) \| \pi_k(s \vert \cdot)) - KL(\pi_*(s \vert \cdot) \| \pi_{k+1}(s \vert \cdot))
    + \ln Z_{s, k}] \\
    &\le \frac{\Ep_{s\sim d^{\pi_*}_\eta}[KL(\pi_*(s \vert \cdot) \| \pi_0(s \vert \cdot))]}{T \ln \beta}
    + \frac{1}{(1-\gamma)T} \sum_{t=0}^{T-1} \Ep[b^{\pi_{k+1}}_\gamma(d^{\pi_*}_\eta) - b^{\pi_k}_\gamma(d^{\pi_*}_\eta)] \\
    &= \frac{\Ep_{s\sim d^{\pi_*}_\eta}[KL(\pi_*(s \vert \cdot) \| \pi_0(s \vert \cdot))]}{T \ln \beta}
    + \frac{\Ep[b^{\pi_T}_\gamma(d^{\pi_*}_\eta) - b^{\pi_0}_\gamma(d^{\pi_*}_\eta])}{(1-\gamma)T} \\
    &\le \frac{\ln A_{\max}}{T \ln \beta}
    + \frac{\Ep[b^{\pi_T}_\gamma(d^{\pi_*}_\eta) - b^{\pi_0}_\gamma(d^{\pi_*}_\eta)]}{(1-\gamma)T} \\
    &\le \frac{\ln A_{\max}}{T \ln \beta}
    + \frac{\Ep[c^{\max}_\gamma(\eta) - c^{\min}_\gamma(\eta)]}{(1-\gamma)T}
\end{align*}
In the final step, we apply \cref{ass:disc-visit-reward}.
Using the fact $b^{\pi_T}_\gamma(\eta) \ge b^{\pi_{T-1}}_\gamma(\eta)$ completes the proof.
\end{proof}
\section{Conclusion}

We give the first proof of convergence for the Natural Policy Gradient algorithm in the infinite-state average-reward queueing setting.
In particular, we demonstrate that by setting the learning rate function $\beta_s$ appropriately,
we can guarantee an $O(1/\sqrt{T})$ convergence rate
to the optimal policy,
as long as the initial policy has a well-behaved relative value function.
Moreover, we demonstrate that in infinite-state average-reward MDPs arising out of queueing theory, the MaxWeight policy satisfies our requirement on the initial policy, allowing our results on the NPG algorithm to apply.

One potential direction for future work would be to tighten the dependence of our convergence rate result on the structural parameters of the MDP, which we did not seek to optimize in this result.
Another potential direction would be to extend our result to an uncountably-infinite-state setting, rather than the countably-infinite setting that we focused on, by tweaking the assumptions made in \cref{sec:assumptions}.

In the reinforcement learning (RL) setting, an important direction for future work would be to build off of our MDP optimization result to prove convergence-rate results for policy-gradient-based RL algorithms in the infinite-state setting.
As important intermediate steps towards this goal,
future work could study the NPG algorithm with function approximation in the infinite-state setting,
as well as computationally-efficient variants of the NPG algorithm,
such as those incorporating neural-network-based policy approximation.

\section{Acknowledgments}

We thank the area editor and the reviewers for their helpful comments and questions, which significantly improved the paper.
We also acknowledge the following sources of funding: AFOSR Grant FA9550-24-1-0002, NSF Grants CNS 23-12714, CCF 22-07547, CNS 21-06801, EPCN-2144316, and CPS-2240982, the Tennenbaum Postdoctoral Fellowship at Georgia Tech, and a Northwestern University startup grant.

\bibliographystyle{plainnat}
\bibliography{refs}

\appendix

\section{Background Lemmas}
\label{app:background}

While this lemma is contained within \cite[Lemma 2]{murthy_convergence_2023}, we reprove it here in a standalone fashion.

\begin{replemma}{lem:monotonic-q}
Policy $\pi_{k+1}$ improves upon policy $\pi_k$ relative to $\pi_k$'s relative-value function $Q_k$:
    \begin{align*}
        Q_k(s, \pi_{k+1}) \ge Q_k(s, \pi_k) = V_k(s)
    \end{align*}
\end{replemma}
\begin{proof}
    Recall the definition of the NPG algorithm:
    \begin{align*}
       \pi_{k+1}(a \mid s) &= \pi_k(a \mid s) \beta_s^{Q_k(s, a)} / Z_{s,k}, 
       \text{where } Z_{s,k} = \sum_{a'} \pi_k(a' \mid s) \beta_s^{Q_k(s, a')}.
    \end{align*}

    Rearranging to solve for $Q_k(s, a)$, we find that
    \begin{align*}
        Q_k(s, a) = \frac{1}{\log \beta_s} \log \left(\frac{Z_{s,k} \pi_{k+1}(a \mid s)}{\pi_k(a \mid s)}\right)
    \end{align*}
    Now, let's start manipulating $Q_k(s, \pi_{k+1})$:
    \begin{align}
        \nonumber
        Q_k(s, \pi_{k+1}) &= \sum_{a} \pi_{k+1}(a \mid s) Q_k(s, a) \\
        &= \sum_a \pi_{k+1}(a \mid s) \frac{1}{\log \beta_s} \log \left(\frac{Z_{s,k} \pi_{k+1}(a \mid s)}{\pi_k(a \mid s)}\right) \\
        \label{eq:kl-term}
        &= \frac{1}{\log \beta_s} \sum_a \pi_{k+1}(a \mid s) \log \left(\frac{\pi_{k+1}(a \mid s)}{\pi_k(a \mid s)}\right)
        + \frac{1}{\log \beta_s} \sum_a \pi_{k+1}(a \mid s) \log Z_{s,k}
    \end{align}
    Note that the left-hand term in \eqref{eq:kl-term} is simply the $KL$ divergence $D_{KL}(\pi_{k+1}(\cdot \mid s) \vert \vert  \pi_k(\cdot \mid s))$.
    As a result, it is positive. Thus,
    \begin{align*}
        Q_k(s, \pi_{k+1}) &\ge \frac{1}{\log \beta_s} \sum_a \pi_{k+1}(a \mid s) \log Z_{s,k} \\
        &= \frac{1}{\log \beta_s} \log Z_{s,k} = \frac{1}{\log \beta_s} \log \sum_{a'} \pi_k(a' \mid s) \beta_s^{Q_k(s, a')}
    \end{align*}
    Note that the log function is concave, so we can apply Jensen's inequality:
    \begin{align*}
        Q_k(s, \pi_{k+1}) &\ge \frac{1}{\log \beta_s} \sum_{a'} \pi_k(a' \mid s) Q_k(s, a') \log \beta_s \\
        &= \sum_{a'} \pi_k(a' \mid s) Q_k(s, a') = Q_k(s, \pi_k) = V_k(s).
        %\qedhere
    \end{align*}
\end{proof}

\section{Bounding relative value}
\label{app:bound-vk-state}

\begin{replemma}{lem:bound-vk-state}
    Given an MDP satisfying \cref{ass:reward},
    and given an initial policy $\pi_0$ such that
    there exist constants $c_0 > 0, c_1 \ge 0$ such that
    \begin{align}
        \label{eq:v-ass-restate}
        V_0(s) \ge -c_0 \rhat_{\max}(s)^2 - c_1,
    \end{align}
    there exists a uniform bound $M_s$:
    \begin{align*}
        M_s := c_2 \vert \rhat_{\max}(s)\vert ^2 + c_3 \vert \rhat_{\max}(s)\vert  + c_4,
    \end{align*}
    for constants $c_2, c_3, c_4 \ge 0$ depending on $c_0, c_1$, and the MDP parameters,
    such that for any NPG iterate $\pi_k$, and any pair of actions $a, a'$,
    \begin{align*}
        Q_k(s, a) - Q_k(s, a') \le M_s.
    \end{align*}
\end{replemma}
\begin{proof}
    The difference $Q_k(s, a) - Q_k(s, a')$ can be separated into the following terms:
    \begin{align*}
        Q_k(s, a) - Q_k(s, a') =
        E_{s' \sim P(s, a)} [V_k(s')]
        - E_{s'' \sim P(s, a')} [V_k(s'')]
        + r(s, a) - r(s, a')
    \end{align*}
    We will bound these terms relative to $\vert \rhat_{\max}(s)\vert $, recalling that rewards are negative.

    Let us start by applying the upper and lower bounds from \cref{lem:bound-vk}.
    Define the following constants, where $z$ is an arbitrary reward threshold below $J_0$:
    \begin{align*}
        c_5 := \frac{J_* - z}{J_0 - z},
        c_6 := \tau^{bound}_{J_0,z} \frac{(J_* - J_0)(c_{\max} - z)(J_* - z)}{(J_0 - z)^2},
        c_7 := \tau^{bound}_{J_0,z} (c_{\max} - J_0) 
    \end{align*}
    Note that $c_5, c_6, c_7$ are all positive.
    Our results hold for all values of $z< J_0$.

    \cref{lem:bound-vk} states that $\min(c_5 V_0(s), V_0(s)) - c_6 \le V_k(s) \le c_7$.
    Our lower bound on $V_0(s)$, \eqref{eq:v-ass-restate}, is negative, so we can simplify the bound to $c_5 \widetilde{V_0}(s) - c_6 \le V_k(s) \le c_7$, where $\widetilde{V_0}$ is our lower bound on $V_0$.
    As a result,
    \begin{align*}
        c_5 E_{s'' \sim P(s, a')} [\widetilde{V_0}(s'')] - c_6 &\le E_{s'' \sim P(s, a')} [V_k(s'')] \\
        E_{s' \sim P(s, a)} [V_k(s')] &\le c_7.
    \end{align*}

    To relate $\widetilde{V_0}(s'')$ to $\rhat_{\max}(s)$,
    we will combine \eqref{eq:v-ass-restate}, our assumption on $V_0(s)$, with \cref{ass:reward}(d):
    \begin{align*}
        \widetilde{V_0}(s'') &\ge -c_0 \rhat_{\max}(s'')^2 - c_1
                  \ge -c_0 (R_3 \rhat_{\max}(s) - R_4)^2 - c_1 \\
                  &= -c_0(R_3 \vert \rhat_{\max}(s)\vert  + R_4)^2 - c_1 \\
        E_{s'' \sim P(s, a')} [V_k(s'')] &\ge -c_0 c_5 (R_3 \vert \rhat_{\max}(s)\vert  + R_4)^2 - c_1 c_5 - c_6
    \end{align*}
    Finally, we can bound $r(s, a) - r(s, a')$ using \cref{ass:reward}(c):
    $r(s, a) - r(s, a') \le R_1 \rhat_{\max}(s)^2 + R_2$.

    Combining these bounds together, we find that
    \begin{align*}
        &Q_k(s, a) - Q_k(s, a')\\
        &\le c_0 c_5 (R_3 \vert \rhat_{\max}(s)\vert  + R_4)^2 + c_1 c_5 + c_6 + c_7 + R_1 \vert \rhat_{\max}(s)\vert ^2 + R_2 \\
        &=  (c_0 c_5 R_3^2 + R_1) \vert \rhat_{\max}(s)\vert ^2 + 2 c_0 c_5 R_3 R_4 \vert \rhat_{\max}(s)\vert  + c_0 c_5 R_4^2 + c_1 c_5 + c_6 + c_7 + R_2 \\
        &= c_2 \vert \rhat_{\max}(s)\vert ^2 + c_3 \vert \rhat_{\max}(s)\vert  + c_4,
    \end{align*}
    where $c_2 := c_0 c_5 R_3^2 + R_1, c_3 := 2 c_0 c_5 R_3 R_4$, and  $c_4 := c_0 c_5 R_4^2 + c_1 c_5 + c_6 + c_7 + R_2$.
    Note that $c_2, c_3,$ and $c_4$ are all nonnegative.
\end{proof}

\section{Queueing Lemmas}

\subsection{Bound on MaxWeight}
\label{app:bound-mw}

This result is proven in \cite[Theorem 4.2.4]{srikant_communication_2013}, for the classic switch model.
For completeness, we now reprove the result for the general GSSE setting.

\begin{replemma}{lem:bound-mw}
    For any GSSE queueing system,
    \begin{align*}
        E_{MW}[f(q(t+1)) - f(q(t)) \mid q(t) = q ] \le c_1 r(q(t)) + c_2,
    \end{align*}
    where $MW$ denotes the MaxWeight policy,
    where $f(q) := \sum_i q^2_i$,
    and where $c_1$ and $c_2$ are the following constants:
    \begin{align*}
        c_1 &= 2 \epsilon \min_i \lambda_i, \qquad c_2 = \ell^2 n
    \end{align*}
\end{replemma}

\begin{proof}

Let $w_i^{MW}(t)$ denote MaxWeight's service on step $t$.
We proceed as follows:
\begin{align*}
    f(q(t+1)) - f(q(t)) &= \sum_i \left( (q_i(t) + v_i(t) - w_i^{MW}(t))^+\right)^2 - \sum_i q_i^2(t) \\
    &\le \sum_i \left( q_i(t) + v_i(t) - w_i^{MW}(t)\right)^2 - \sum_i q_i^2(t) \\
    &= \sum_i (2 q_i(t) + v_i(t) - w_i^{MW}(t))(v_i(t) - w_i^{MW}(t)) \\
    &= \sum_i 2 q_i(t) (v_i(t) - w_i^{MW}(t)) + \sum_i (v_i(t) - w_i^{MW}(t))^2
\end{align*}
Note that the second term is bounded.
Recall that $\ell$ is the maximum number of jobs that can arrive or depart from a class in one time step.
As a result, $(v_i(t) - w_i^{MW}(t))^2 \le \ell^2$, so the second summation is bounded by $\ell^2 n$.

Taking expectations, we find that
\begin{align}
    \label{eq:mid-lyapunov}
    E[f(q(t+1)) - f(q(t)) \mid q(t)=q] \le 2 \sum_i q_i (\lambda_i - \mu_i^{MW}) + \ell^2 n
\end{align}

Let us now lower bound $\sum_i q_i \mu_i^{MW}$.
Recall that Maxweight selects the service option $j$
which maximizes $\sum_i q_i \mu_i^j$.
Thus, for all $j$,
\begin{align}
    \label{eq:mw-ineq}
    \sum_i q_i \mu_i^{MW} \ge \sum_i q_i \mu_i^j.
\end{align}

Recall that we assumed that $\lambda$ lies within the stability region. In particular, by \eqref{eq:convex}, there exists a vector $\gamma \ge 0$ of service-option weights such that
\begin{align*}
    \sum_j \gamma_j = 1 \text{ and } \forall i, (1+\epsilon)\lambda_i \le \sum_j \gamma_j \mu_i^j
\end{align*}

Let us multiply this inequality by $q_i$ and sum over $i$. We then find that
\begin{align}
    \label{eq:comp1}
    \sum_{i,j} q_i \gamma_j \mu_i^j \ge (1+\epsilon) \sum_i q_i \lambda_i.
\end{align}

On the other hand, let us take the MaxWeight inequality \eqref{eq:mw-ineq},
multiply by $\gamma_j$ and sum over $j$.
We then find that
\begin{align}
    \nonumber
    \sum_i q_i \mu_i^{MW} \sum_j \gamma_j &\ge \sum_{i,j} q_i \gamma_j \mu_i^j \\
    \label{eq:comp2}
    \sum_i q_i \mu_i^{MW} &\ge \sum_{i,j} q_i \gamma_j \mu_i^j
\end{align}

Combining \eqref{eq:comp1} and \eqref{eq:comp2}, we find the desired lower bound on $\sum_i q_i \mu_i^{MW}$: 
\begin{align}
    \label{eq:mw_bound}
    \sum_i q_i \mu_i^{MW} \ge (1+\epsilon) \sum_i q_i \lambda_i
\end{align}

Substituting this bound into \eqref{eq:mid-lyapunov}, we find that
\begin{align*}
    E[f(q(t+1)) - f(q(t)) \mid q(t)=q] \le - 2 \epsilon \sum_i q_i \lambda_i + \ell^2 n
\end{align*}

Recall that $r(q) = -\sum_i q_i$.
Note that $\sum_i q_i \lambda_i \ge -r(q) \min_i \lambda_i$.
As a result,
taking $c_1 = 2 \epsilon \min_i \lambda_i$ and taking $c_2 = n$,
the proof is complete.

\end{proof}

The above argument straightforwardly generalizes to the setting where the reward is the $\alpha$-th moment of the total queue length, using the Lyapunov function $f(q) = \sum_i q^{\alpha+1}$.

\subsection{Verifying \cref{ass:reward}(d) for $\alpha$-moment reward}
\label{app:verify}

We want to verify \cref{ass:reward}(d) for the GSSE MDP with reward $-(\sum_i q_i)^\alpha$, for $\alpha \ge 1$.

\begin{lemma}
    In any GSSE MDP with the $\alpha$-moment reward function,
    there exists a pair of constants $R_3 \ge 1$ and $R_4 \ge 0$ such that
    for any pair of states $s, s'$ such that there is a nonzero probability of transitioning from $s$ to $s'$
    under some action $a$,
    \begin{align*}
        \rhat_{\max}(s') \ge R_3 \rhat_{\max}(s) - R_4
    \end{align*}
\end{lemma}
\begin{proof}
Let $q$ be the total queue length of state $s$.
Note that the total queue length of states $s'$ is at most $q+\ell n$, because at most $\ell$ jobs of each of the $n$ job classes can arrive
in a given time step.
Let $n' \le \ell n$ denote this maximum possible number of arriving jobs.

We have
\begin{align*}
    \vert r_{\max}(s)\vert  = q^\alpha, \quad \vert r_{\max}(s')\vert  \le (q+n')^\alpha
\end{align*}

Let us set $R_3 = 2$. We want to find some $R_4$ such that

\begin{align*}
    (q+n')^\alpha \le 2 q^\alpha + R_4, \quad \forall n', \alpha \ge 1
\end{align*}

First, let's apply the mean value theorem to the quantity $(q+n')^\alpha - q^\alpha$.
The function $x^\alpha$ has derivative $\alpha x^{\alpha - 1}$, which is increasing for positive $x$.
Thus,
\begin{align*}
    (q+n)^\alpha - q^\alpha \le n' \alpha (q+n')^{\alpha - 1}
\end{align*}
We will bound $n' \alpha (q+n')^{\alpha - 1}$ in two parts:
For large $q$, we will show $n' \alpha (q+n')^{\alpha - 1} \le q^\alpha$.
For small $q$, we will select $R_4$ such that $n' \alpha (q+n')^{\alpha - 1} \le R_4$.
Our split between these bounds is the value $q = 2 n' \alpha$,
where $W$ is the Lambert $W$ function.

First, for large $q$,
\begin{align*}
    \forall q \ge 2n'\alpha,\quad 
    q^\alpha &\ge 2n'\alpha q^{\alpha-1} \\
    &= 2n'\alpha (q+n')^{\alpha-1} \left(\frac{q+n'}{q}\right)^{-\alpha + 1} \\
    &= 2n'\alpha (q+n')^{\alpha-1} \left(1+\frac{n'}{q}\right)^{-\alpha + 1} \\
    &\ge 2n'\alpha (q+n')^{\alpha-1} \left(1+\frac{n'}{2n'\alpha}\right)^{-\alpha + 1} \\
    &\ge 2n'\alpha (q+n')^{\alpha-1} \left(1+\frac{1}{2\alpha}\right)^{-\alpha + 1} \\
    &\ge 2n'\alpha (q+n')^{\alpha-1} e^{-1/2} \\
    &\ge n'\alpha (q+n')^{\alpha-1}
\end{align*}
To cover small $q$, we simply set
\begin{align*}
    R_4 = n' \alpha (2\alpha n' + n')^{\alpha - 1} \ge n' \alpha (q+n')^{\alpha - 1} \quad \forall q \le 2n'\alpha
\end{align*}

Now we can conclude that 
\begin{align*}
    \forall q, \quad n' \alpha (q+n')^{\alpha - 1} &\le q^\alpha + R_4 \\
    (q+n')^\alpha - q^\alpha &\le q^\alpha + R_4 \\
    (q+n')^\alpha &\le 2 q^\alpha + R_4
\end{align*}
\end{proof}
\end{document}